\documentclass{article}

\usepackage{times}
\usepackage{soul}
\usepackage{url}
\usepackage{hyperref}
\usepackage[small]{caption}
\usepackage{amsmath}
\usepackage{amsthm}
\usepackage[switch]{lineno}
\usepackage{longtable}
\usepackage{supertabular}
\usepackage{multirow}

\usepackage[in]{fullpage}  
\usepackage{arxiv} 
\usepackage{mathpazo} 

\usepackage[utf8]{inputenc} 
\usepackage[T1]{fontenc}    
\usepackage{url}            
\usepackage{booktabs}       
\usepackage{amsfonts}       
\usepackage{microtype}      
\usepackage{lipsum}         

\usepackage{graphicx}  
\usepackage{float} 
\usepackage{subfigure} 
\usepackage{algorithm}
\usepackage{algorithmic}
\usepackage{tabulary}

\usepackage{amssymb}
\usepackage{pifont}

\usepackage{hyperref}       %
\usepackage{url}            %
\usepackage{booktabs}       %
\usepackage{amsfonts}       %
\usepackage{nicefrac}       %
\usepackage{microtype}      %
\usepackage{xcolor}         %
\usepackage{graphicx}
\usepackage{enumitem}
\usepackage{subfigure}
\usepackage{diagbox}

\def\gA{{\mathcal{A}}}

\def\gD{{\mathcal{D}}}

\def\gG{{\mathcal{G}}}

\def\gM{{\mathcal{M}}}

\def\gP{{\mathcal{P}}}

\def\gR{{\mathcal{R}}}
\def\gS{{\mathcal{S}}}

\newcommand{\benchname}{ImagineBench}

\hypersetup{
    colorlinks=false,
    citecolor=black,
    linkcolor=black,
    citecolor=black
}

\title{ImagineBench: Evaluating Reinforcement Learning with Large Language Model Rollouts}

\usepackage{authblk}
\author{%
  Jing-Cheng Pang*\textsuperscript{\rm 1,2}, 
  Kaiyuan Li*\textsuperscript{\rm 1,2}, 
  Yidi Wang*\textsuperscript{\rm 1,2}, 
  Si-Hang Yang\textsuperscript{\rm 1,2}, 
  Shengyi Jiang\textsuperscript{\rm 3}, 
  Yang Yu\textsuperscript{\rm 1,2,$\diamond$}\\
  \textsuperscript{\rm 1} National Key Laboratory for Novel Software Technology, Nanjing University, China \\ \& School of Artificial Intelligence, Nanjing University, China \\
  \textsuperscript{\rm 2}Polixir.ai \\
  \textsuperscript{\rm 3}The University of Hong Kong \\
  \textsuperscript{*} Equal contribution\\
  \textsuperscript{$\diamond$} Corresponding: yuy@nju.edu.cn
}

\date{}
\begin{document}

\maketitle

\begin{abstract}
    A central challenge in reinforcement learning (RL) is its dependence on extensive real-world interaction data to learn task-specific policies. While recent work demonstrates that large language models (LLMs) can mitigate this limitation by generating synthetic experience (noted as \textit{imaginary rollouts}) for mastering novel tasks, progress in this emerging field is hindered due to the lack of a standard benchmark. To bridge this gap, we introduce \benchname, the first comprehensive benchmark for evaluating offline RL algorithms that leverage both real rollouts and LLM-imaginary rollouts. The key features of \benchname~include: (1) datasets comprising environment-collected and LLM-imaginary rollouts; (2) diverse domains of environments covering locomotion, robotic manipulation, and navigation tasks; and (3) natural language task instructions with varying complexity levels to facilitate language-conditioned policy learning. Through systematic evaluation of state-of-the-art offline RL algorithms, we observe that simply applying existing offline RL algorithms leads to suboptimal performance on unseen tasks, achieving 35.44\% success rate in hard tasks in contrast to 64.37\% of method training on real rollouts for hard tasks. This result highlights the need for algorithm advancements to better leverage LLM-imaginary rollouts. Additionally, we identify key opportunities for future research: including better utilization of imaginary rollouts, fast online adaptation and continual learning, and extension to multi-modal tasks. Our code is publicly available at \url{https://github.com/LAMDA-RL/ImagineBench}.
\end{abstract}

\section{Introduction}
\label{sec:intro}

\begin{figure}[t]
	\centering
	\includegraphics[width=\linewidth]{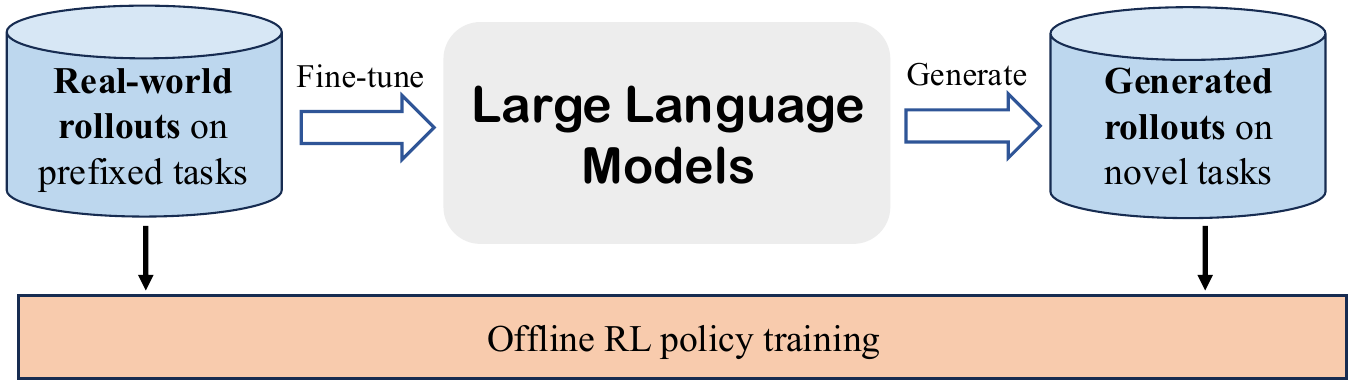}
	\caption{We benchmark the problem of RL with LLM-imaginary rollouts. The LLM is fine-tuned to generate imaginary rollouts, followed by RL policy training using real and imaginary rollouts.}
	\label{sec:illustration_rlim}
\end{figure}

Developing knowledgeable agents capable of mastering diverse, unseen tasks represents a critical frontier in artificial intelligence. While reinforcement learning (RL) provides a framework for training such agents \cite{alphago,dqn,alphastar}, its reliance on extensive real-world interaction data remains a fundamental bottleneck, especially for generalizing to unseen tasks. In contrast, humans excel at acquiring new skills through mental imagination, enabling skill acquisition without direct physical interaction. 
Recent advances in large language models (LLMs) have demonstrated their potential to replicate this capability \cite{kalm,learning_from_book} by leveraging their extensive knowledge to generate synthetic task execution trajectories. \textit{A promising and effective way to leverage LLMs is to generate imaginary rollouts}, referring to synthetic experience that simulates the task execution trajectories, and then apply offline RL algorithms to train the policy \cite{kalm,learning_from_book}. 
We formalize this methodology under the framework of \textit{\underline{RL} from \underline{Im}aginary Rollouts} (RLIM). RLIM fine-tunes an LLM on environment interaction data and then prompts it to generate synthetic rollouts for novel tasks (see Fig. \ref{sec:illustration_rlim}).

Though preliminary successes in robotics manipulation \cite{kalm}, football playing \cite{learning_from_book} and browser automation \cite{agenttrek} highlight RLIM’s promise, previous works often utilize their customized environments and varied LLM architectures. The reported performance improvements may not reliably demonstrate effective utilization of LLM knowledge due to inconsistent evaluation protocols. As a result, progress in this field could be hindered.

To address this gap, we introduce \benchname, the first comprehensive benchmark designed to systematically evaluate offline RL algorithms that train policy with both real rollouts and LLM-imaginary rollouts. \benchname~have three key features: 
(1) It provides datasets that include both real rollouts collected from the environment and imaginary rollouts generated by the fine-tuned LLMs. These datasets could reduce the computational costs of developing new RLIM algorithms. 
(2) It covers diverse domains, including locomotion, robotic manipulation, and navigation.
(3) It provides natural language instruction paired with the rollouts, which are divided into various difficulty levels. This feature aligns with the recent advances in building instruction-following agents \cite{talar,saycan,RLC}.
Through extensive empirical analysis of state-of-the-art offline RL algorithms, we demonstrate that while naively combining real and imaginary rollouts generally improves performance on unseen tasks, there is still a clear gap between the current score and the performance of training with real rollouts (35.44\% and 64.37\%). This gap underscores the need for novel algorithms to leverage LLM-generated rollouts better.

Our contributions are as follows: 
We implement and open-source the \benchname, the first comprehensive benchmark for RLIM, providing standardized environments, datasets, and evaluation protocols.
Additionally, we reveal the limitations of existing offline RL methods when applied to hybrid real-imaginary datasets.
Lastly, we identify key challenges and opportunities as future directions, including offline RL algorithm development for better utilizing LLM-imaginary rollouts, fast online adaptation and continual learning, and extending RLIM to multi-modal tasks.

\section{Related Work}
\label{sec:related_work}

\textbf{RL with LLM-imaginary rollouts.} Recent advances in leveraging the general knowledge of LLMs to build knowledgeable agents for interactive and physical tasks have established a promising research frontier \cite{kalm}. The central challenge lies in that LLMs can not directly handle numerical control signals for decision-making tasks \cite{kalm,LLM_robot}. To address this, researchers have explored using LLMs to generate imaginary decision-making rollouts that are then used for RL policy training. For instance, KALM \cite{kalm} fine-tunes LLMs to produce low-level control rollouts, which are then used to train RL policies via offline RL algorithms. This approach demonstrates how domain-specific knowledge embedded in LLMs can be effectively distilled to handle novel tasks. Similarly, URI \cite{learning_from_book} employs LLMs to generate control trajectories by prompting them with instructional texts from tutorial books, enabling policy training without environmental interaction. AgentTrek \cite{agenttrek} extends this paradigm to browser automation by synthesizing task execution rollouts at scale, followed by imitation learning to train the agent. Beyond low-level control, InCLET \cite{inclet} introduces a framework where LLMs generate textual imaginary rollouts, enhancing the agent’s ability to interpret natural language instructions and derive task representations.
While these studies highlight the potential of LLM-imaginary rollouts for building knowledgeable agents, existing work primarily focuses on algorithmic development. In contrast, \benchname~introduces the first comprehensive benchmark to systematically evaluate the performance, generalizability, and limitations of LLM-driven imaginary rollout methods.

\textbf{Existing benchmarks in RL}. The rapid development of RL has given rise to a diverse array of benchmarks. These benchmarks fall into three primary categories: online, offline, and off-dynamics, each handling challenges within specific training paradigms.
Benchmarks have emerged to address distinct challenges across training paradigms. Online training benchmarks, such as OpenAI Gym \cite{gym}, MuJoCo \cite{mujoco}, and the DeepMind Control Suite \cite{dmc}, have long served as foundational tools for evaluating agents that learn through direct environmental interaction, emphasizing exploration and sample efficiency in dynamic settings like Atari 2600 games \cite{atari} and continuous control tasks. Meanwhile, the rise of offline RL promotes the development of benchmarks like NeoRL \cite{neorl}, D4RL \cite{d4rl} and RL Unplugged \cite{rl_unplugged}, which contain large-scale, pre-collected datasets to evaluate agents’ ability to learn from static data while mitigating distributional shift and extrapolation errors in domains ranging from robotic manipulation to locomotion. Besides, off-dynamics benchmarks, including ODRL \cite{odrl} and Meta-World ML1 \cite{meta_world}, evaluate generalization under shifts in dynamics, such as altered physical parameters or visual perturbations, challenging agents to adapt policies to unseen environmental conditions. Generally, these benchmarks reflect the evolving demands of RL research, from sample-efficient online exploration to robust offline learning and cross-domain adaptability, providing standardized frameworks to evaluate the strengths and limitations of algorithms across diverse environments. 
In contrast, \benchname~is the first benchmark specifically designed to evaluate how effectively RL algorithms that utilizes LLM-imaginary rollouts to training the policy, offering scenarios to measure the benefits and limitations of LLM-driven knowledge transfer in decision-making tasks.

\textbf{Offline reinforcement learning.}
This work considers utilizing offline RL algorithms to train the policy. Offline RL \cite{offlinerlsurvey,bcq} enables agents to learn effective policies from static datasets without online environment interactions. Early approaches to offline RL, such as BCQ \cite{bcq} and BEAR \cite{bear}, addressed distributional shift by constraining learned policies to remain close to the behavior policy through explicit policy regularization or uncertainty-based action clipping. Subsequent advances introduced CQL \cite{cql}, which penalizes Q-value overestimation for out-of-distribution actions, and implicit constraint methods like TD3+BC \cite{td3_bc} that balance policy improvement with behavior cloning. Decision transformer \cite{decision_trans} has also explored leveraging trajectory-level optimization via sequence modelling.
Despite these advancements, offline RL remains constrained by dataset quality: policies trained on narrow or non-diverse data often fail in unseen scenarios. Model-based RL \cite{mbrl_survey} addresses this by learning a dynamics model from offline data, enabling policy optimization through simulated rollouts. Methods like MOPO \cite{mopo} and MOReL \cite{morel} incorporate uncertainty quantification to construct pessimistic models, mitigating model bias and distributional mismatch.  
These environment models are typically learned from scratch because they can not leverage prior knowledge about the world, which may result in rollouts that do not align with real-world data distributions.

\section{Background}
\label{sec:background}

\textbf{Reinforcement learning.} We consider RL problem where the agent completes natural language instructions. The environment can be modeled as a goal-augmented Markov Decision Process \cite{sutton2011reinforcement,o3f}, represented by the tuple $\gM = \left( \gS, \gA, \gP, \gR, \gamma, \gG \right)$, where $\gS$, $\gA$ denote the state space and action space, respectively. $\gP$ denotes transition function of the environment, $\gR$ the reward function that evaluates the agent's behavior, $\gamma$ the discount factor, and $\gG$ the set of natural language goals. 
The objective of RL is to find a policy $\pi: \gS \times \gG \rightarrow \Delta (\gA)$ that maximize the culmulative reward: $J(\pi)=\mathbb{E}_\pi [\sum_{t=0}^{\infty}\gamma^t r(s_t,a_t)]$.
In this work, we focus on environments with structured, vectorized state spaces, where each dimension encodes interpretable, domain-specific features.
We call the state and action data collected from the environment as the \emph{real environmental rollouts}, and the rollouts generated by LLMs as \emph{imaginary rollouts}.

\textbf{Offline reinforcement learning with LLM-imaginary rollouts.} Traditional offline RL focuses on offline policy training from a static environmental dataset. 
In this paper, we consider RL with both real and LLM-imaginary rollouts. Formally, consider we have (1) a real dataset $\gD$ collected from the real environment, and (2) a LLM-imaginary datasets\footnote{We will elaborate on how LLMs are trained to generate the rollouts in Sec. \ref{sec:data_collection}.} $\gD^I$, which is generated by LLMs. Both real and imaginary datasets consist of paired language goals and corresponding decision-making rollouts: $\{G^k, (s_0^k,a_0^k,s_1^k,a_1^k,\cdots) \}_{k=1}^K$. Here, the sequence $(s_0^k,a_0^k,s_1^k,a_1^k,\cdots)$ represents a rollout of states and actions $(s_i^k,a_i^k)$ to complete the goal $G^k$. The primary objective is to find a policy that achieves high rewards on unseen goal distributions, represented as $\gG'$. We elaborate on the `unseen goal distributions' in Sec. \ref{sec:task_level}.
\section{ImagineBench Details}
\label{sec:dataset}

\begin{figure}[t]
    \centering
    \includegraphics[width=1\linewidth]{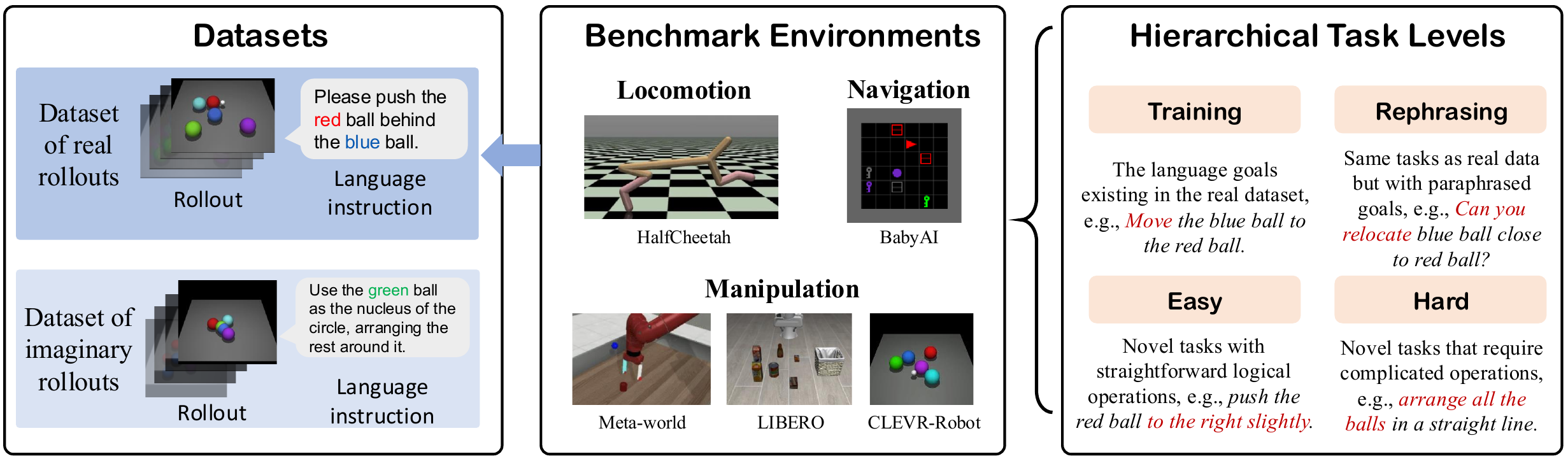}
    \caption{Overview of \benchname. Three key features of our benchmark: (1) datasets of both real and LLM-imaginary rollouts, (2) diverse domains of environments, and (3) natural language instructions divided into various task levels. Examples shown in the `Datasets' panel are from CLEVR-Robot environment.}
    \label{fig:imaginary_rollout_generation}
\end{figure}

\benchname~involves a wide range of decision-making environments, including locomotion, manipulation and navigation. For each environment, \benchname~provides two datasets as illustrated in Sec. \ref{sec:background}: a dataset of real rollouts collected from the environments, and a dataset of imaginary rollouts generated by LLM. We will briefly introduce the benchmark environments in Sec. \ref{sec:bench_env} and how the datasets are constructed in Sec. \ref{sec:data_collection}. Last, Sec. \ref{sec:task_level} introduces the definition of different levels of task complexity.

\subsection{Benchmark Environments}
\label{sec:bench_env}

\begin{figure}[t]
    \centering
    \includegraphics[width=1\linewidth]{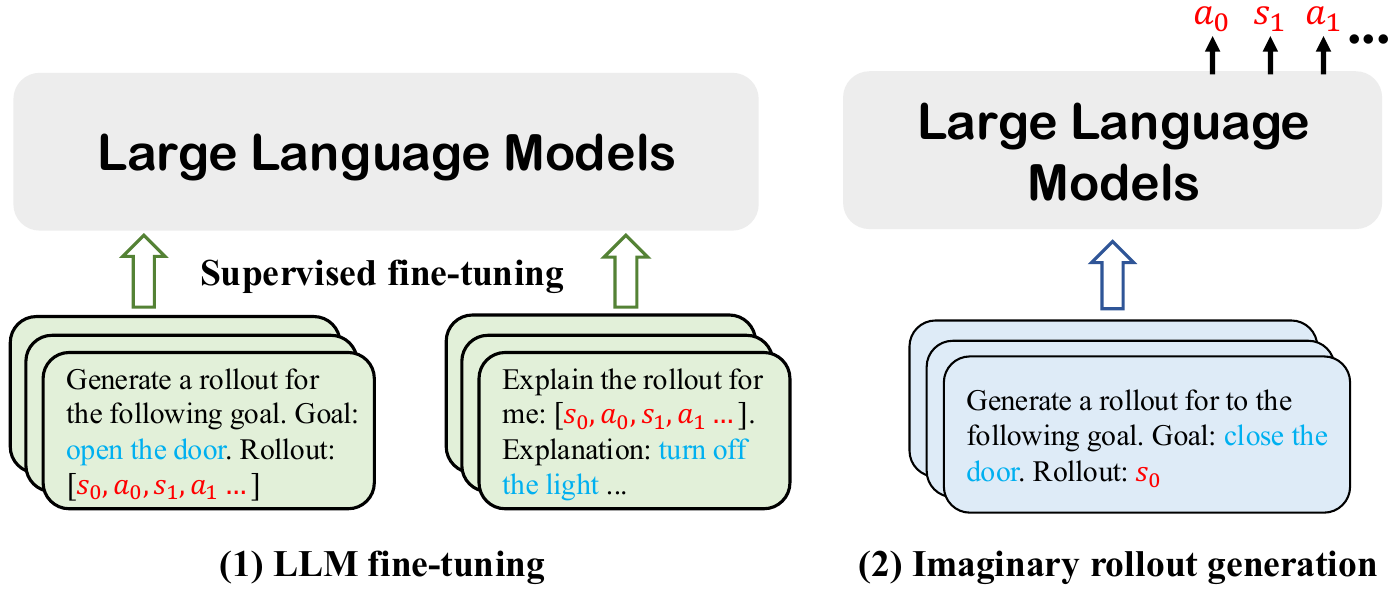}
    \caption{Illustration of the generation of LLM-imaginary rollouts. The LLM is first fine-tuned with the environment data, and then prompted to generate the rollouts for novel tasks.}
    \label{fig:overall_framework}
\end{figure}

The Benchmark Environment panel in Fig. \ref{fig:overall_framework} shows the visualization of the environments used in \benchname. We present the environment statistics in Tab. \ref{tab:dataset_statistics}. 

\textbf{Meta-world} \cite{meta_world} agent controls a Sawyer robot to manipulate various objects, e.g., doors, drawers and windows. In novel tasks, the agent requires to manipulate assuming that there is a wall in front of the object.
The state space is $\mathbb{R}^{91}$, encodeing the robot’s joint angles and object positions/orientations, while the action space is $\mathbb{R}^{4}$, controlling the gripper's movement and open/close. The reward function combines task (or sub-task) completion signals with a negative distance metric between the gripper and target location.

\textbf{CLEVR-Robot} \cite{clevr_robot} environment requires the agent to manipulate five colored balls to reach a target configuration. The state space is $\mathbb{R}^{10}$, encoding the positions of five balls, with action space of 40-dimensional discrete actions, using one-hot vectors to specify directional movement for individual ball. The reward is calculated as reduction in distance between the current state and the target configuration compared to the previous step, adding a terminal reward for task completion.

\textbf{BabyAI} \cite{babyai} is a gridworld environment, which modifies the original environment's language-conditioned navigation tasks with full observability. The state space is $\mathbb{R}^{17}$, encoding object positions (agent, keys, doors, balls) using absolute grid coordinates and RGB attributes. The action space comprises 7-dimensional discrete movement primitives (left/right/up/down) and object interactions (pickup/drop/toggle). The rewards are calculated as the shortest-path distance to the goal object, plus a sparse completion reward.

\textbf{LIBERO} \cite{liu2023libero} controls a robot arm to complete various manipulation tasks. LIBERO originally consists of four task suites, each containing 10 tasks. \benchname~uses LIBERO-Object suite and additionally designs novel tasks such as sequential-pick-and-place. The state space is $\mathbb{R}^{44}$, representing the joint position and object position/poses, while action space of $\mathbb{R}^{7}$ specifying joint angle deltas for arm movement and gripper open/close. Similar to Meta-world, we provide distance-based reward to guide the agent reach the target object, and terminal judgment when a sub-task or the entire task is completed as the final step reward. 

\textbf{MuJoCo} \cite{mujoco} is a physics-based simulation platform widely used for continuous control tasks in reinforcement learning. In our case, \benchname~uses the HalfCheetah robot. The state ($\mathbb{R}^{18}$) consists of positional values and velocities of different joints, while the action space ($\mathbb{R}^{6}$) represents the torques applied on 6 robot joints. The reward function combines forward velocity toward the target direction with control efficiency (minimizing joint torque costs).

\newcolumntype{C}[1]{>{\centering\arraybackslash}p{#1}}

\begin{table}[h]
    \centering
    \small
    \begin{tabular}{C{2.5cm}|C{4cm}|C{1.4cm}|C{1.2cm}|C{1.8cm}|C{2.2cm}}
    \toprule
     & \textbf{Meta-world} & \textbf{CLEVR-Robot} & \textbf{BabyAI} & \textbf{LIBERO} & \textbf{MuJoCo}  \\   \toprule
    Observation space & $\mathbb{R}^{91}$ & $\mathbb{R}^{10}$ & $\mathbb{Z}^{17}$ & $\mathbb{R}^{44}$ & $\mathbb{R}^{18}$ \\ \midrule
    Action space & $\mathbb{R}^{4}$ & Discrete (40) & Discrete (7) & $\mathbb{R}^{7}$ & $\mathbb{R}^{6}$ \\ \midrule
    \# of real rollout & 20,000 & 100,000 & 19,200 & 29,780 & 16,000 \\ \midrule
    \# of imaginary rollout (Rephrase) & 72,400 & 72,400 & 19,200 & 12,000 & 10,000 \\ \midrule
    \# of imaginary rollout (Easy) & 8,000 & 5,600 & 18,000 & 24,000 & 6,000 \\ \midrule
    \# of imaginary rollout (hard) & 4,000 & 1,680 & 18,000 & 1,3000 & 9,000 \\ \midrule
    Training task & Reach, Push, Pick-place, Button-press, Door-unlock, Door open, Window-open, Faucet-open, Coffee-push, Coffee-button-press & Move & Goto, Pickup, Open, Put-next & Pick, Place  & Run-forward, Run-backward, Jump-forward, Jump-backward \\ \midrule
    Rephrasing task & \multicolumn{5}{c}{Same as training (with rephrasing instructions)}  \\ \midrule
    Easy task & Reach-wall, Push-wall, Pick-place-wall, Button-press-wall, Door-lock, Door-close, Window-close, Faucet-close & One-step-move & Open-go, Open-pick, Go-wall, Go-center & Pick-and-place, Pick-and-place-to-unseen, Reach & Run-forward-faster, Run-backward-faster \\ \midrule
    Hard task & Make-coffee, Locked-door-open, Hammer, Soccer & Sequential-move, Make-line, Put-pile, Make-circle & Open-lock, Put-line, Put-pile & Sequential-pick-and-place, Pick-and-place-aside, Pick-out & Run-forward-then-backward, Run-backward-then-forward, Jump-in-place \\
    \bottomrule
    \end{tabular}
    \caption{Statistics overview of environments. Refer to Sec. \ref{sec:task_level} for /*details of `Training, Rephrasing, Easy and Hard' tasks.}
    \label{tab:dataset_statistics}
\end{table}

\subsection{Dataset Collection}
\label{sec:data_collection}

The dataset collection procedure consists of two steps: (1) \textit{Real rollout collection} from the environment. In this step, we first obtain an expert policy that can complete the given tasks with a high success rate, and then use the expert policy to collect rollouts in the environment. Meanwhile, a rollout is labelled with a natural language instruction when collected. (2) \textit{Imaginary rollout collection} from LLM. In this step, the LLM is fine-tuned on the rollout-instruction pairs from the environment, and then prompted to generate rollouts for novel tasks-.

\textbf{Real rollout collection.}
To collect real rollouts, we first obtain an expert policy specific to each environment and then use the policy to collect rollouts:
\begin{itemize}[leftmargin=0.5cm]
    \item Meta-world \& CLRVR-Robot: Use a pre-collected offline dataset of 100,000 rollout-goal pairs, each comprising state, action, and environment-built-in reward sequences for completing natural language goals.
    \item BabyAI: Employ a rule-based policy to generate 19,200 rollout-goal pairs, with rewards based on agent-target distance.
    \item LIBERO: Apply behavior cloning to public LIBERO datasets to train a policy, yielding 30,000 rollout-goal pairs with object-target distance rewards.
    \item MuJoCo: Train an expert policy online using the SAC algorithm to collect 16,000 rollout-goal pairs.
\end{itemize}

All real rollouts are annotated with natural language instructions during collection.

\textbf{Imaginary rollout collection.}
Fig. \ref{fig:imaginary_rollout_generation} presents the process of fine-tuning LLM to generate imaginary rollouts\footnote{In \benchname, the imaginary rollouts are generated by the Llama-2-7b-chat-hf \cite{llama2} model.}. 
To enable LLM to generate synthetic task-specific rollouts, we first fine-tune them on real rollout-instruction pairs. The objective of this step is to enable LLM to interpret the meaning of states, actions, dynamics and rollouts of the given environment. Following \cite{kalm}, we fine-tune the LLM using the dataset to perform three different tasks via supervised fine-tuning (SFT), and model the LLM grounding problem as an instruction-following problem since the LLM demonstrates excellent performance following given natural language instructions to generate a desired answer. The SFT paradigms include:

\begin{itemize}[leftmargin=0.5cm]
    \item Dynamics prediction: The LLM predicts environmental dynamics changes. Given the current state $s_t$ and action $a_t$, the LLM predicts the subsequent state.
    \item Rollout explanation: The LLM is presented with a rollout sequence $s_0, a_0, s_1,\cdots$, and it is required to describe the rollout with natural language.
    \item Rollout generation: The LLM generates a rollout that aligns with a specified goal $G$. \label{pattern:rg}
\end{itemize}

Since LLMs are initially trained on textual data and can not handle numerical data, we use a pre-trained LLM as the backbone model and modify it with additional layers to handle environmental data. This framework can be easily extended to visual observation tasks and different LLMs by integrating appropriate neural network architectures.

Then we employ the fine-tuned LLM to generate imaginary rollouts given the initial state $s_0$ and the goal: $\{a_0,s_1,a_1,\cdots\}\leftarrow \gM(GOP,s_0)$. Here, $GOP$ stands for \emph{goal-oriented prompt}: ``Generate a rollout for the following goal: [GOAL]. Rollout:”, where ``[GOAL]” is a placeholder for various goals that reflect different skills.

\subsection{Hierarchical Task Levels}
\label{sec:task_level}

\benchname~defines hierarchical task levels indicating various levels of tasks.

\begin{itemize}[leftmargin=0.5cm]
    \item \textbf{Training}: The instructions appeared in the real dataset. Including training tasks is to evaluate whether the policy preserves the ability to perform these seen tasks.
    \item \textbf{Rephrasing}: The agent performs the same tasks as real data but receives paraphrased instructions which are not present in the data. For example, the goal in offline data is \emph{move the blue ball to the front of the red ball}, while the paraphrased goal could be \emph{I really dislike how the red ball is positioned in front of the blue ball. Could you exchange their places?}
    \item \textbf{Easy}: The agent is tasked with different manipulation tasks that do not exist in the dataset, requiring the agent to generalize to easy, unseen tasks.
    \item \textbf{Hard}: The agent faces tasks substantially different from those in the offline dataset, which require a complex composition of behaviors, such as ``Gather all balls together'', and ``Move five balls to a straight line'' in the CLEVR-Robot environment. 
\end{itemize}

\textbf{Evaluation protocols.} \benchname~utilizes success rate as the evaluation metrics. For each task in the benchmark, \benchname~offers the evaluation function to judge whether the task has been successful, e.g., 5cm positional accuracy for robotic manipulation, or 85\% semantic consistency for HalfCheetah.
\section{Experiment}
\label{sec:experiment}

In this section, we conduct experiments to address three key questions regarding \benchname: (1) How is the quality of the imaginary rollouts (Sec. \ref{sec:analysis_rollout})? (2) How are the benchmark results of popular offline RL algorithms on \benchname~(Sec. \ref{sec:benchmark_result})? and (3) How does the method perform if trained with real rollouts rather than imaginary rollouts on novel tasks  (Sec. \ref{sec:performance_with_real_novel})? We first introduce the experimental setting.

\subsection{Experiment Setting}

\textbf{Baselines.} We give benchmark results with several representative offline RL methods that train policies using static datasets. The methods include:
(1) Behavior Cloning (\textbf{BC}), a supervised learning baseline that directly imitates actions from the dataset.
(2) Conservative Q-Learning (\textbf{CQL}) \cite{cql}, which learns a conservative Q-function to prevent the policy from overestimating expected returns.
(3) \textbf{BCQ} \cite{bcq}, which employs perturbation networks to generate conservative policy updates near offline data.
(4) \textbf{TD3+BC} \cite{td3_bc}, which combines TD3’s (\cite{td3}) stability with BC constraints to enforce similarity to demonstrated behavior.
(5) \textbf{PRDC} \cite{prdc}, which uses a tree-search method to regularize the policy toward the nearest state-action pairs in the offline data.
(6) \textbf{COMBO} \cite{combo}, which uses ensemble environment models to enforce uncertainty-aware policy learning.
(7) Soft Actor-Critic (\textbf{SAC}) \cite{sac}, which is originally an online RL algorithm, can be applied in the offline setting for comparison.

Due to the varying application scope of different algorithms, we evaluate algorithms (BC, CQL, BCQ, TD3+BC, PRDC, COMBO) on MuJoCo, LIBERO and Meta-world, and algorithms (BC, BCQ, CQL, SAC) on CLEVR-Robot and BabyAI. \textit{We use `w/ IR' to represent the methods trained with imaginary rollouts.}

\textbf{Implementation details.} All baseline methods are implemented using the OfflineRL framework \cite{offlinerl_polixir} and d3rlpy library \cite{d3rlpy}, two widely-adopted codebases. Policy optimization relies on the Adam algorithm \cite{kingma2017adam}. Performance metrics are averaged across results from the final five training checkpoints to ensure stability. Unless otherwise specified, baselines encode natural language instructions using BERT \cite{bert}, and concatenate the language encoding with the environment observation.
Offline RL training employs three random seeds to validate robustness. Each training batch uniformly samples equal proportions of data from the real and LLM-imaginary datasets. All experiments are executed on 64 AMD EPYC 9374F 32-core processors, 8 NVIDIA GeForce RTX 4090 GPUs, and 1TB RAM to facilitate parallelized computation.

\subsection{Analysis on LLM-Imaginary Rollouts}
\label{sec:analysis_rollout}

We investigate the quality of the LLM-imaginary rollouts from three aspects: (1) consistency that evaluates the matching rate between generated rollouts and the given goal; (2) transition correctness that describes whether LLM generates unreal data for one-step transition, i.e., the agent moves more than two steps at one time; (3) dynamics legality that measures whether LLM generates illegal states, e.g., two agent exists at the same time. 

\begin{table}[h]
	\centering
	\begin{tabular}{c|c|c|c|c}
         \toprule
		  & Rephrasing & Easy & Hard  \\ \toprule
		Consistency  & 88.0 &  43.8 & 25.8 \\ \midrule
		Transition  & 96.0 & 82.2 & 72.9 \\ \midrule
		Legality  & 98.5 & 81.1 & 66.8 \\ \bottomrule
	\end{tabular}
	\caption{Statistical results on the quality of LLM-imaginary rollouts. The reported results are the LLM-imaginary rollouts for BabyAI environment.}
	\label{tab:rollout_quality}
\end{table}

Tab. \ref{tab:rollout_quality} presents the statistical results about the quality of LLM-imaginary rollouts on BabyAI environments. The results show that the imaginary rollouts for rephrasing goals are highly quality. This is because the LLM, fine-tuned on prefixed goals, can generalize to these rephrasing goals very well. As for Hard goals, the consistency between the novel goal and the imaginary rollouts gets lower. However, the transition correctness and dynamics legality ratios are still higher than 65\%, indicating the LLM-imaginary rollouts basically adhere to environmental dynamics.

\begin{figure}[h]
    \centering
    \includegraphics[width=1\linewidth]{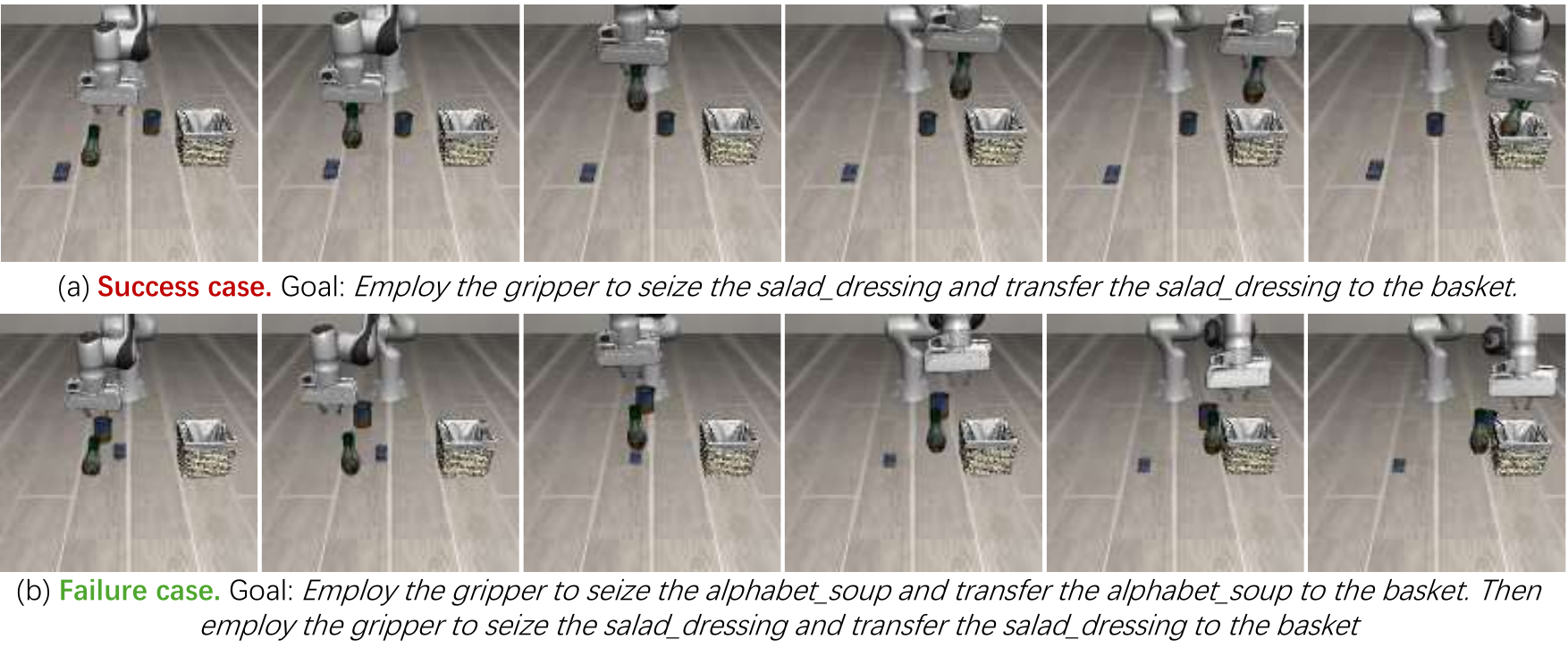}
    \caption{Examples of the LLM-imaginary rollouts for novel goals. The figures are obtained by rendering the states in LLM-imaginary rollouts.}
    \label{fig:imagine_example}
\end{figure}

\textbf{Examples of the LLM-imaginary rollouts.} Tab. \ref{tab:rollout_quality} presents the ratios of the generated rollouts that are consistent with the goals. To further investigate the quality of the generated rollouts, we showcase illustrative examples of the imaginary rollouts in Fig. \ref{fig:imagine_example}. We reset the environment to the generated state to obtain the visualization image. We observe that the generated rollouts can generally reflect the given goals. For example, the robot is conducting the object manipulation as required. However, there are still some mismatches when the goal becomes complicated (pick A, then pick B), where the LLM may generate wrong trajectories (e.g., simultaneous picking instead of sequential execution, as shown in the failure case in Fig. \ref{fig:imagine_example}). Even so, the imaginary rollouts demonstrate a tendency for managing the novel goals, which are never seen by the agent before.

\subsection{Benchmark Result}
\label{sec:benchmark_result}

\begin{figure}[t]
\centering
\rotatebox{90}{\indent \ \indent \ \indent MuJoCo}
\subfigure{
\includegraphics[width=0.225\textwidth]{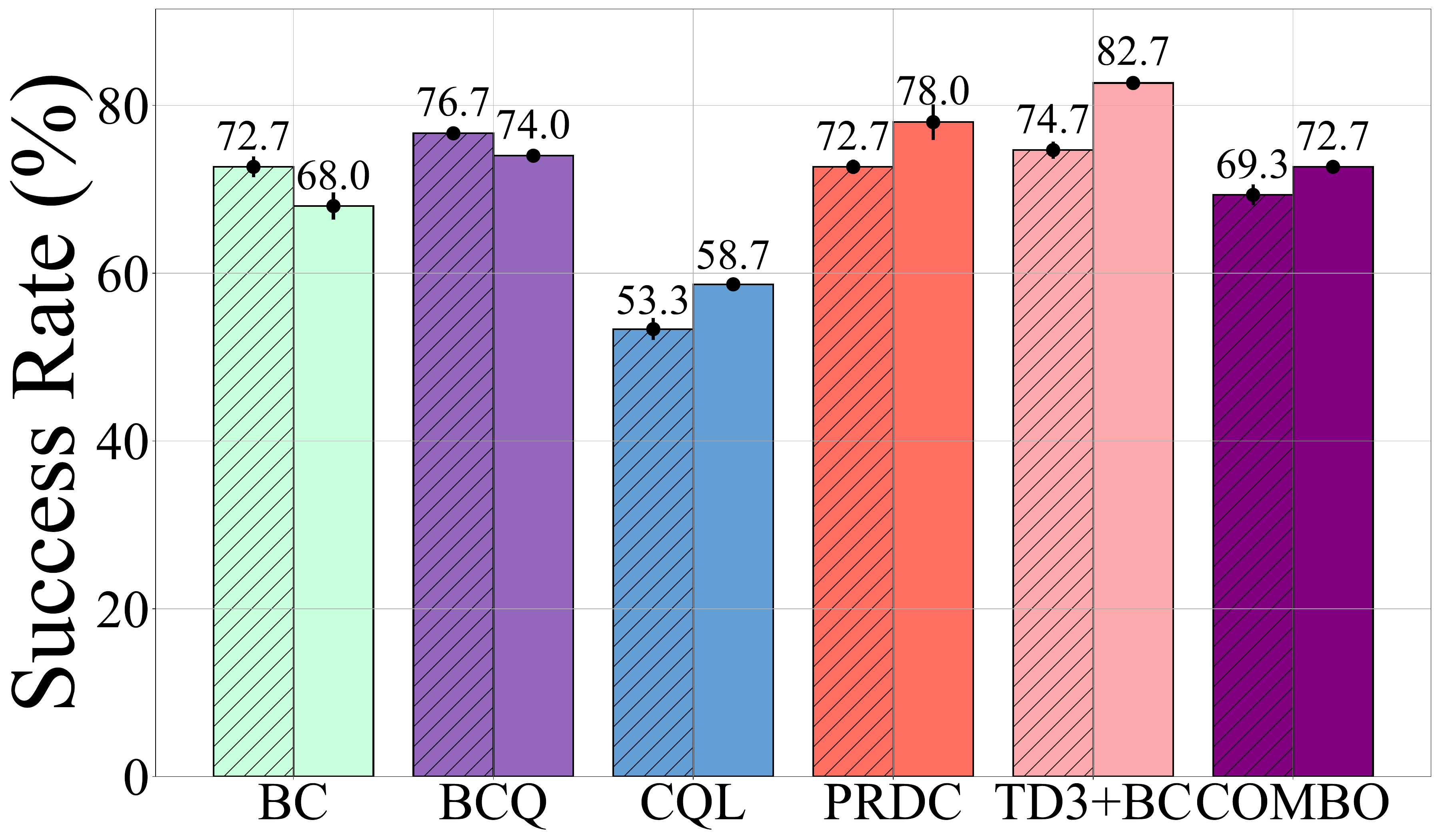}
}
\subfigure{
\includegraphics[width=0.225\textwidth]{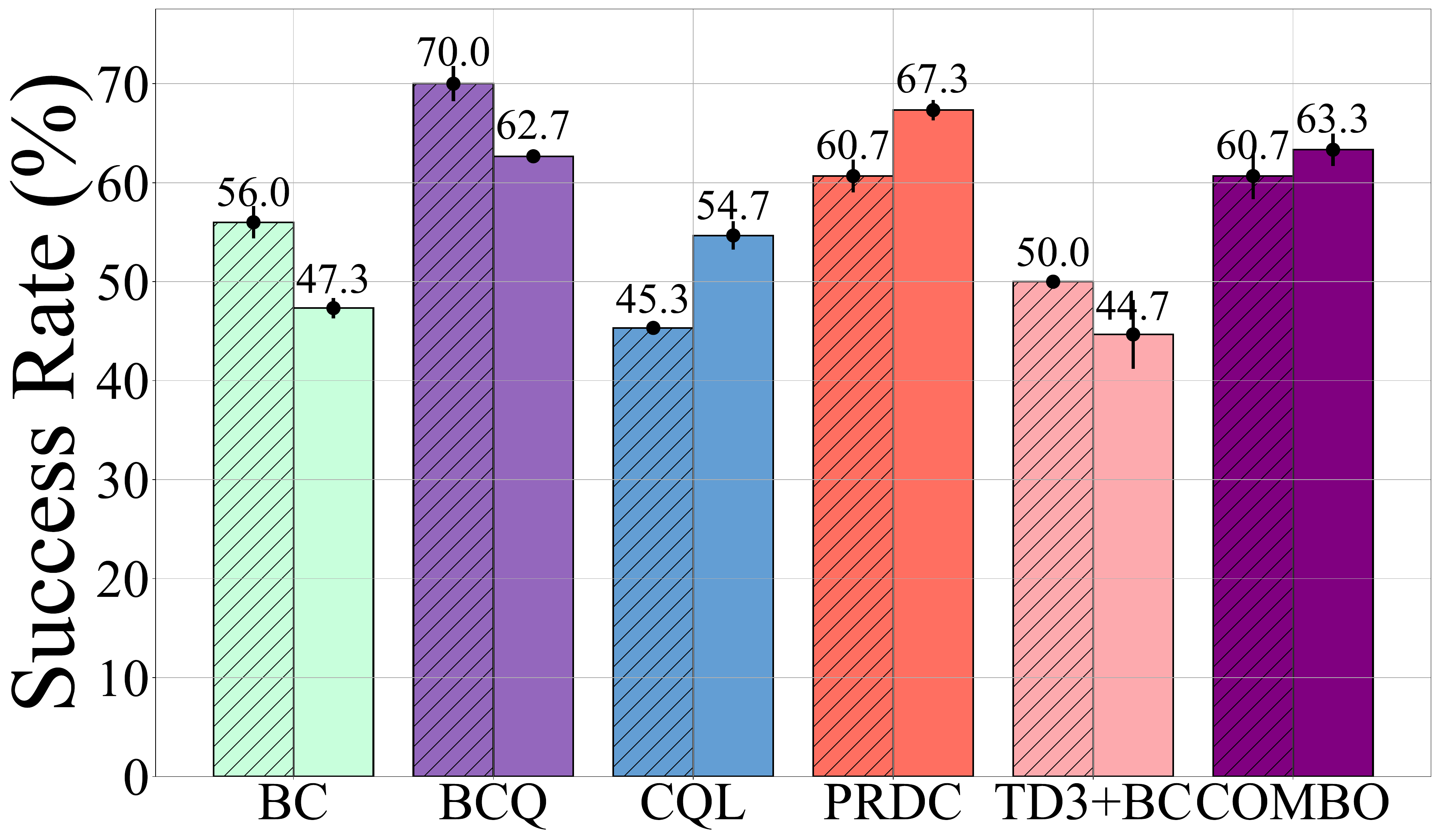}
}
\subfigure{
\includegraphics[width=0.225\textwidth]{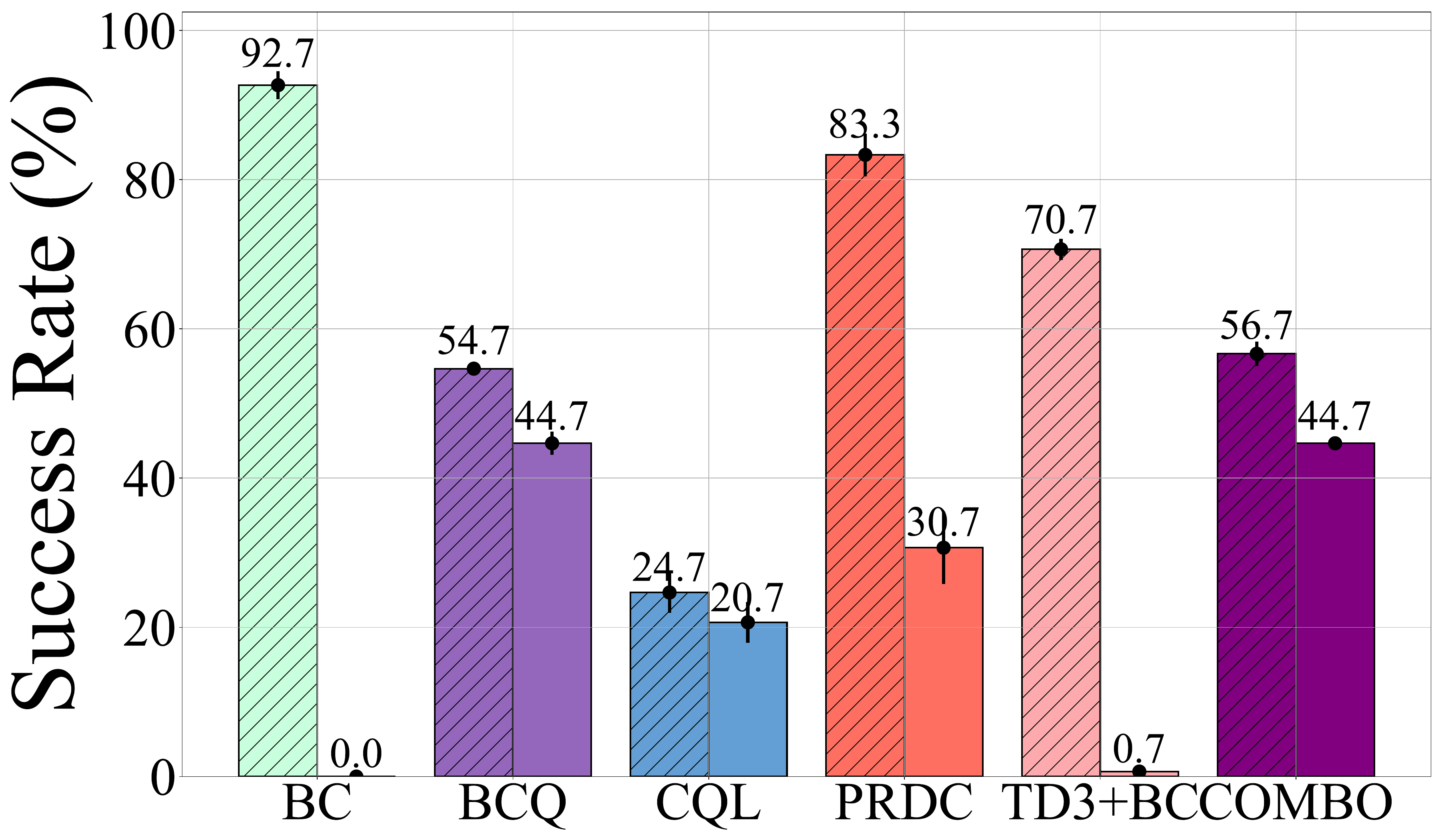}
}
\subfigure{
\includegraphics[width=0.225\textwidth]{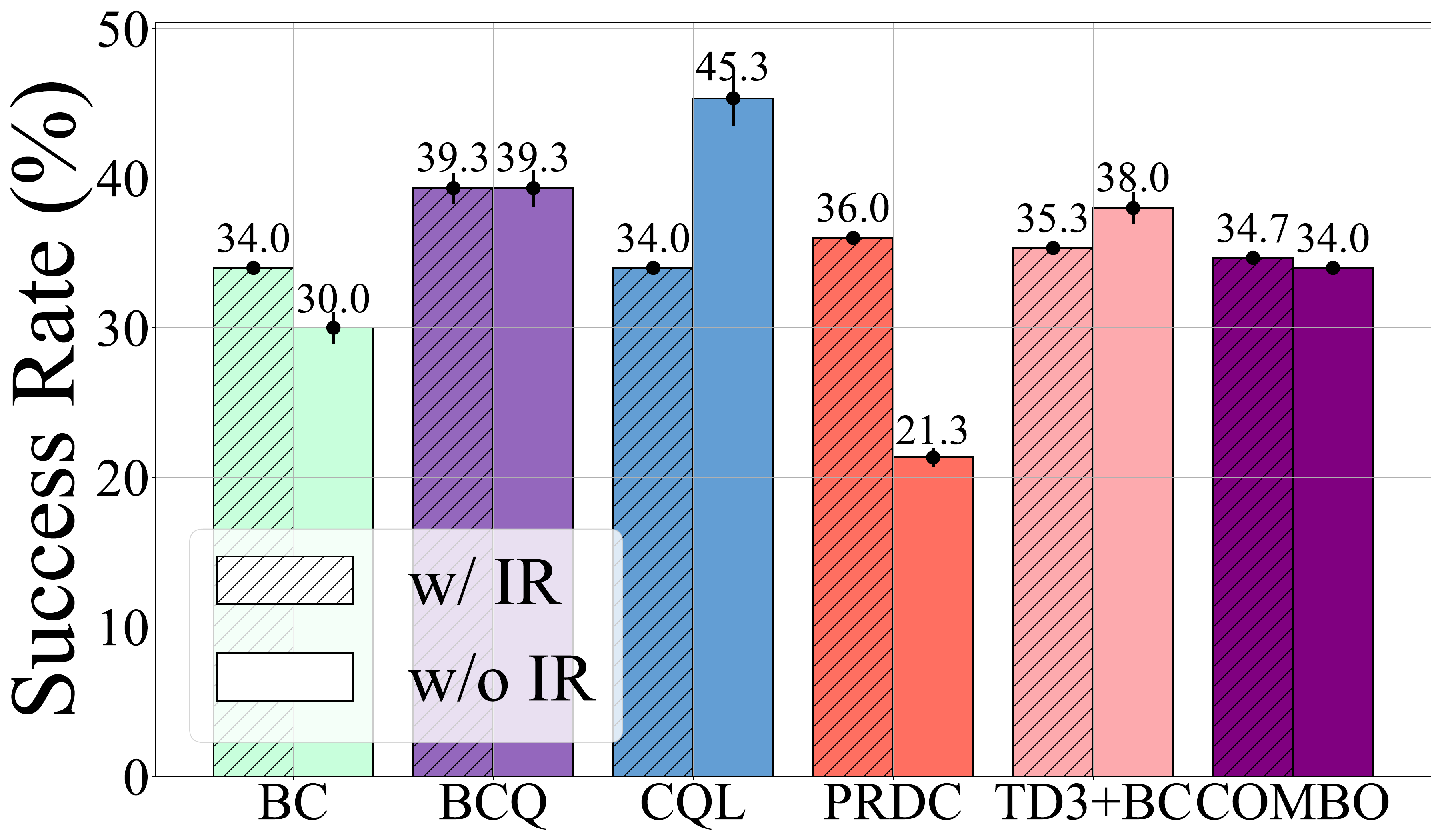}
} \\
\rotatebox{90}{\indent \ \indent \ \indent LIBERO}
\subfigure{
\includegraphics[width=0.225\textwidth]{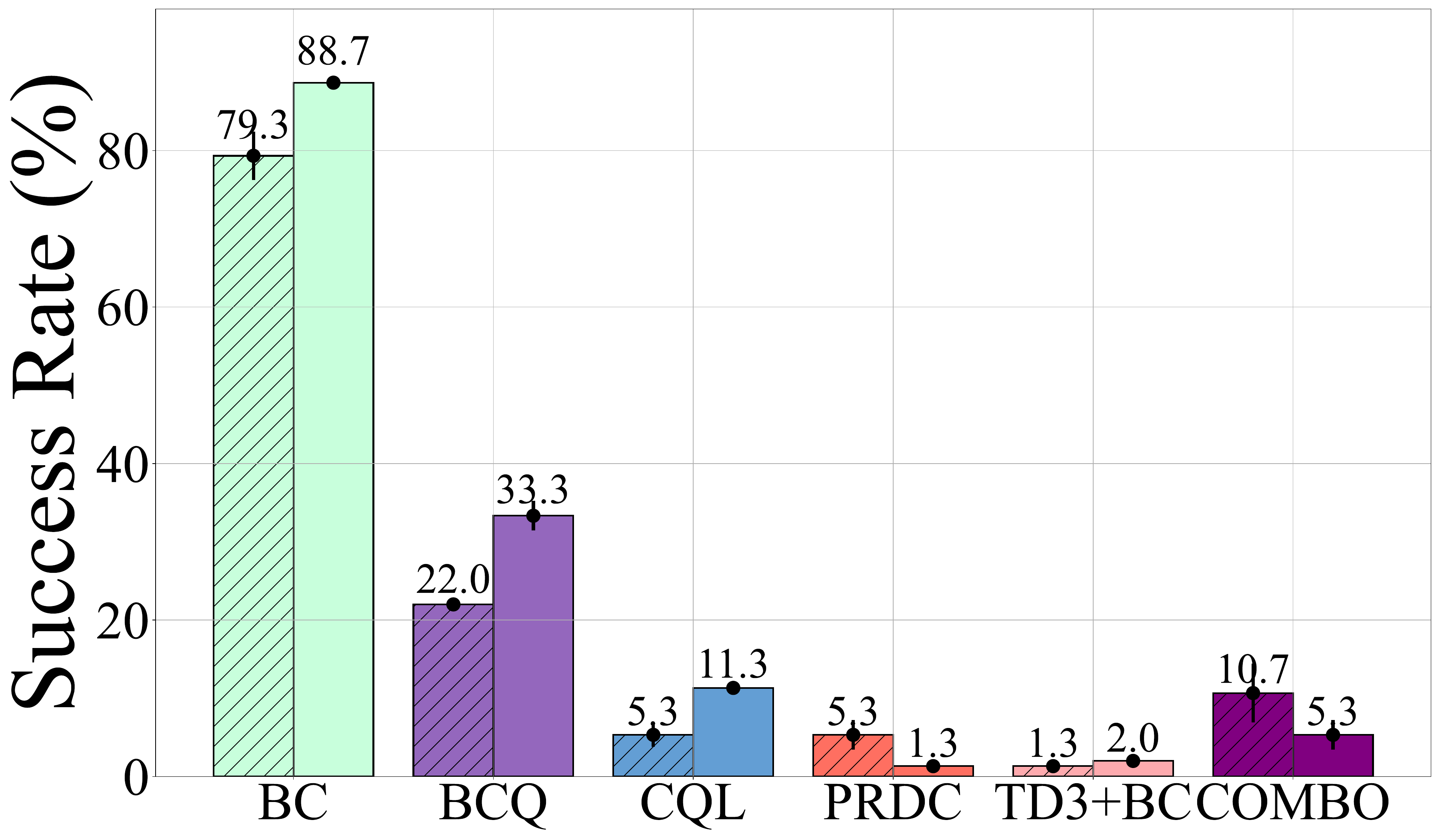}
}
\subfigure{
\includegraphics[width=0.225\textwidth]{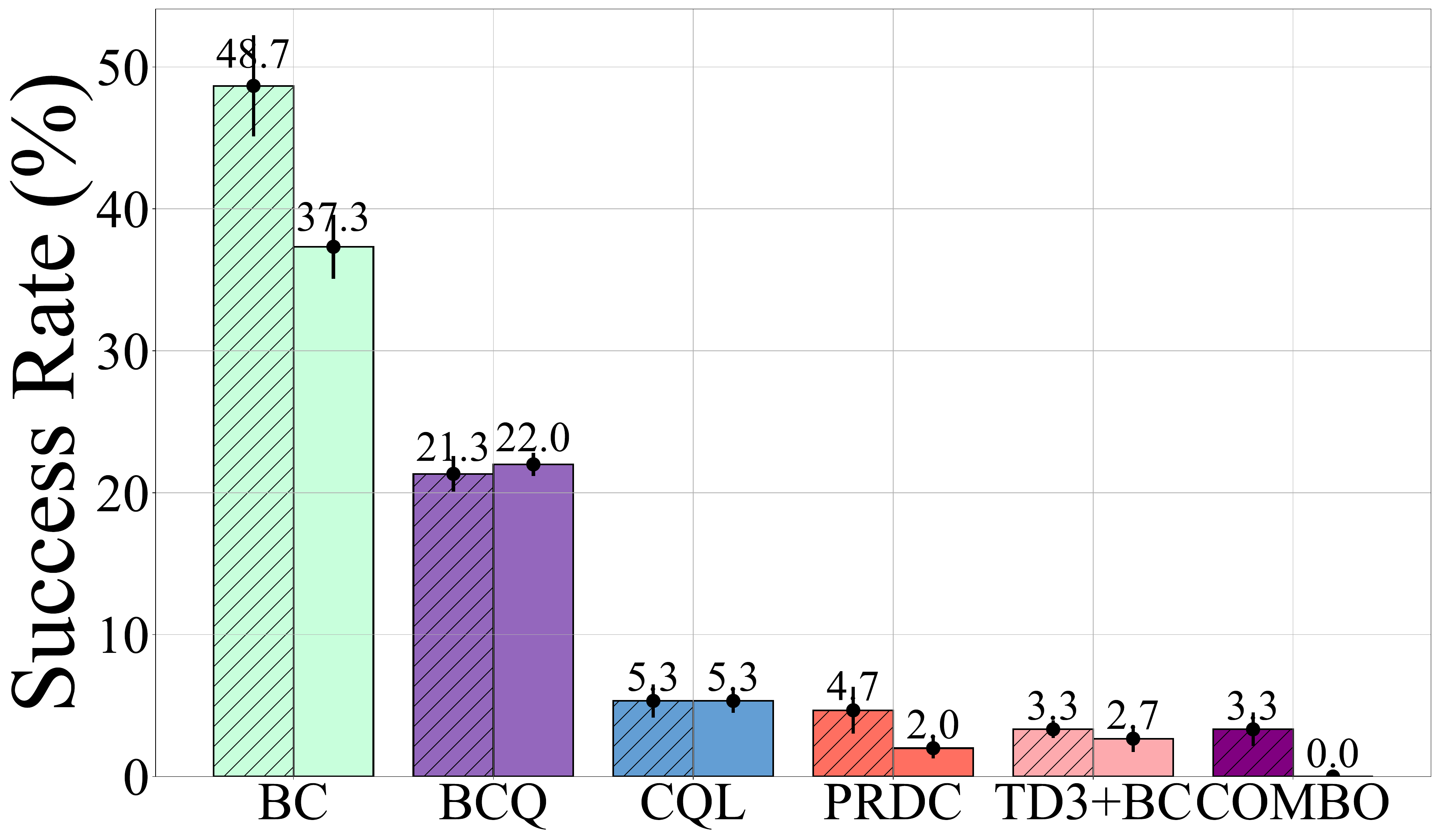}
}
\subfigure{
\includegraphics[width=0.225\textwidth]{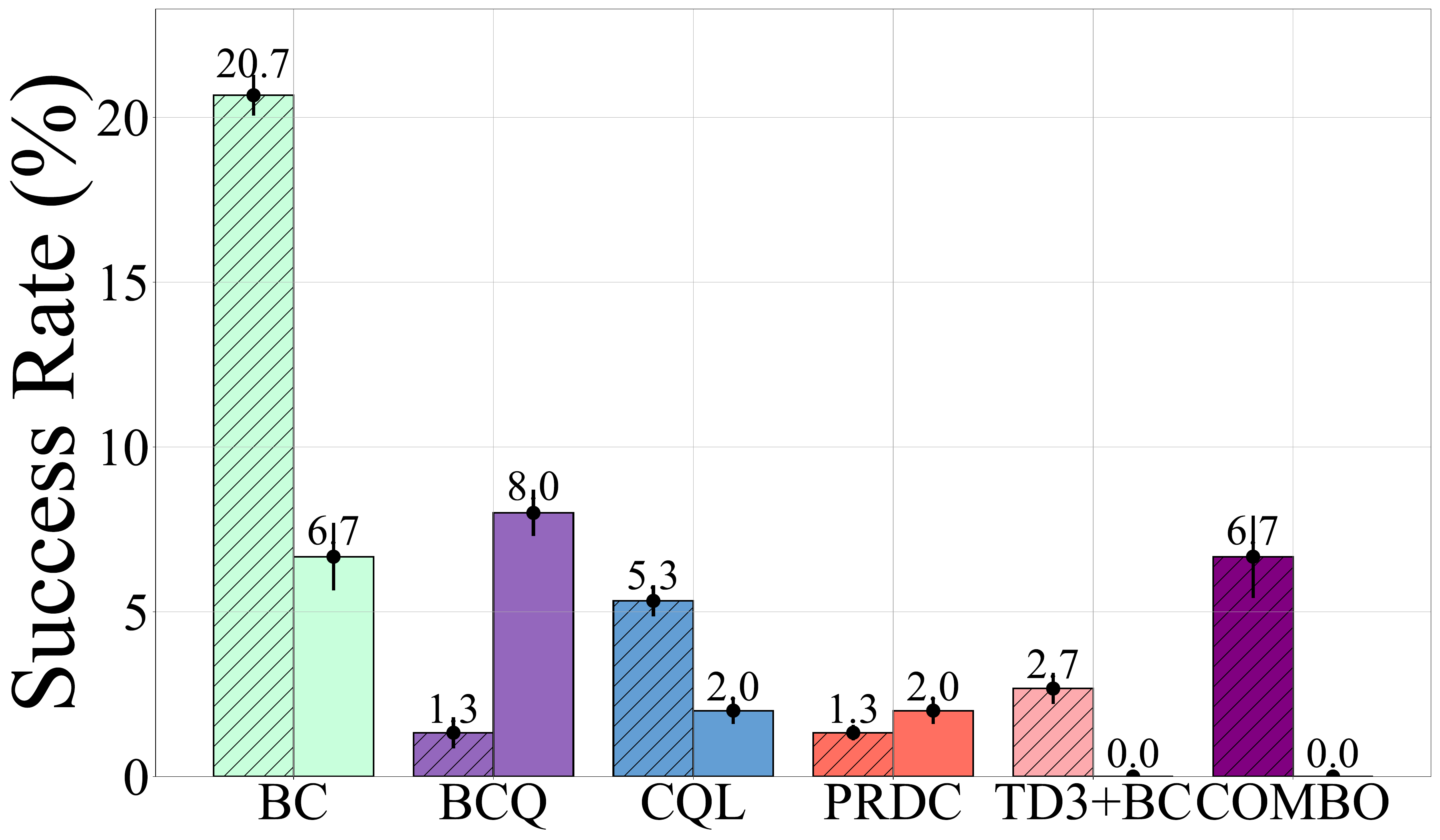}
}
\subfigure{
\includegraphics[width=0.225\textwidth]{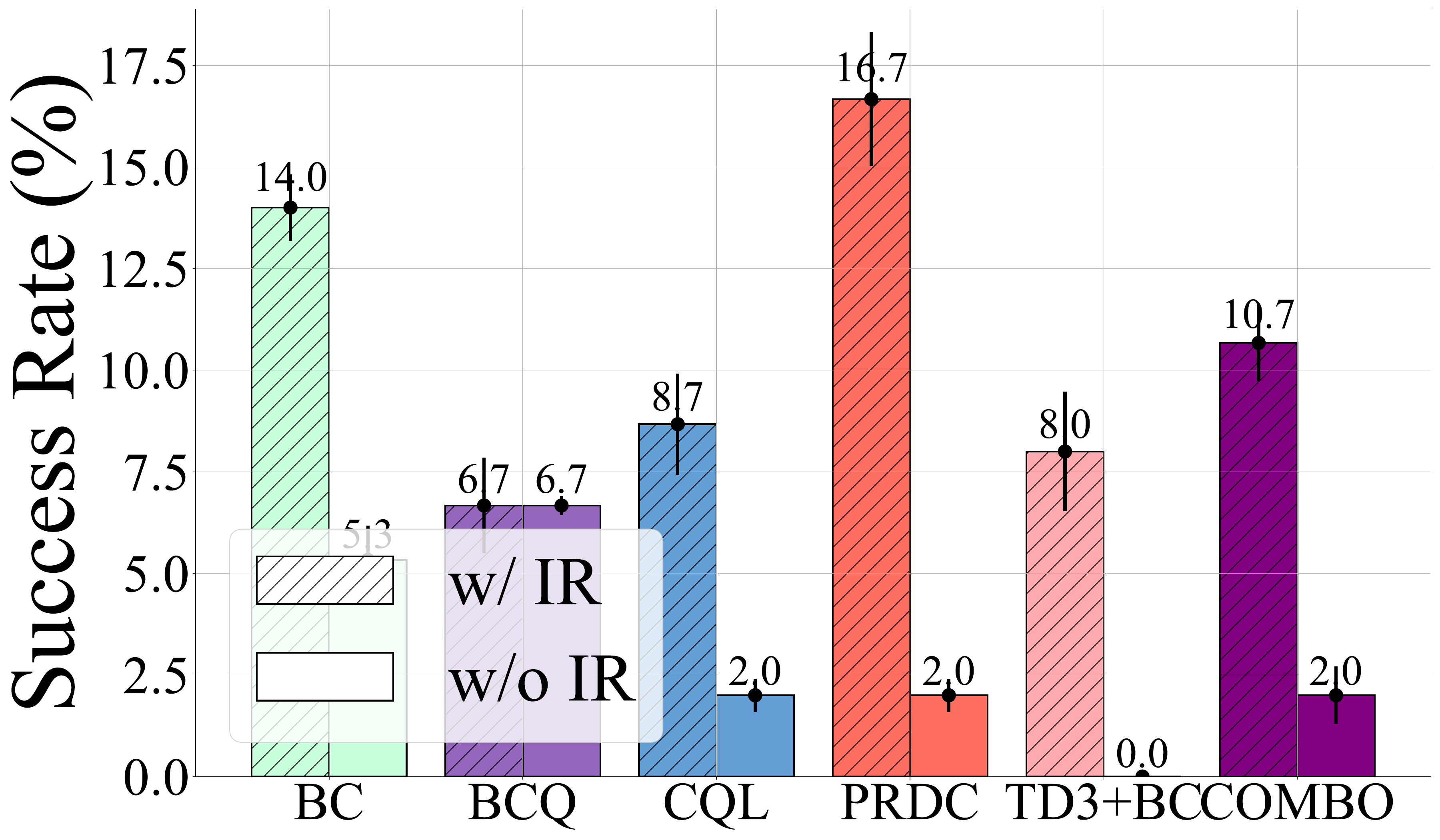}
} \\
\rotatebox{90}{\indent \ \indent \ \indent Meta-world}
\subfigure{
\includegraphics[width=0.225\textwidth]{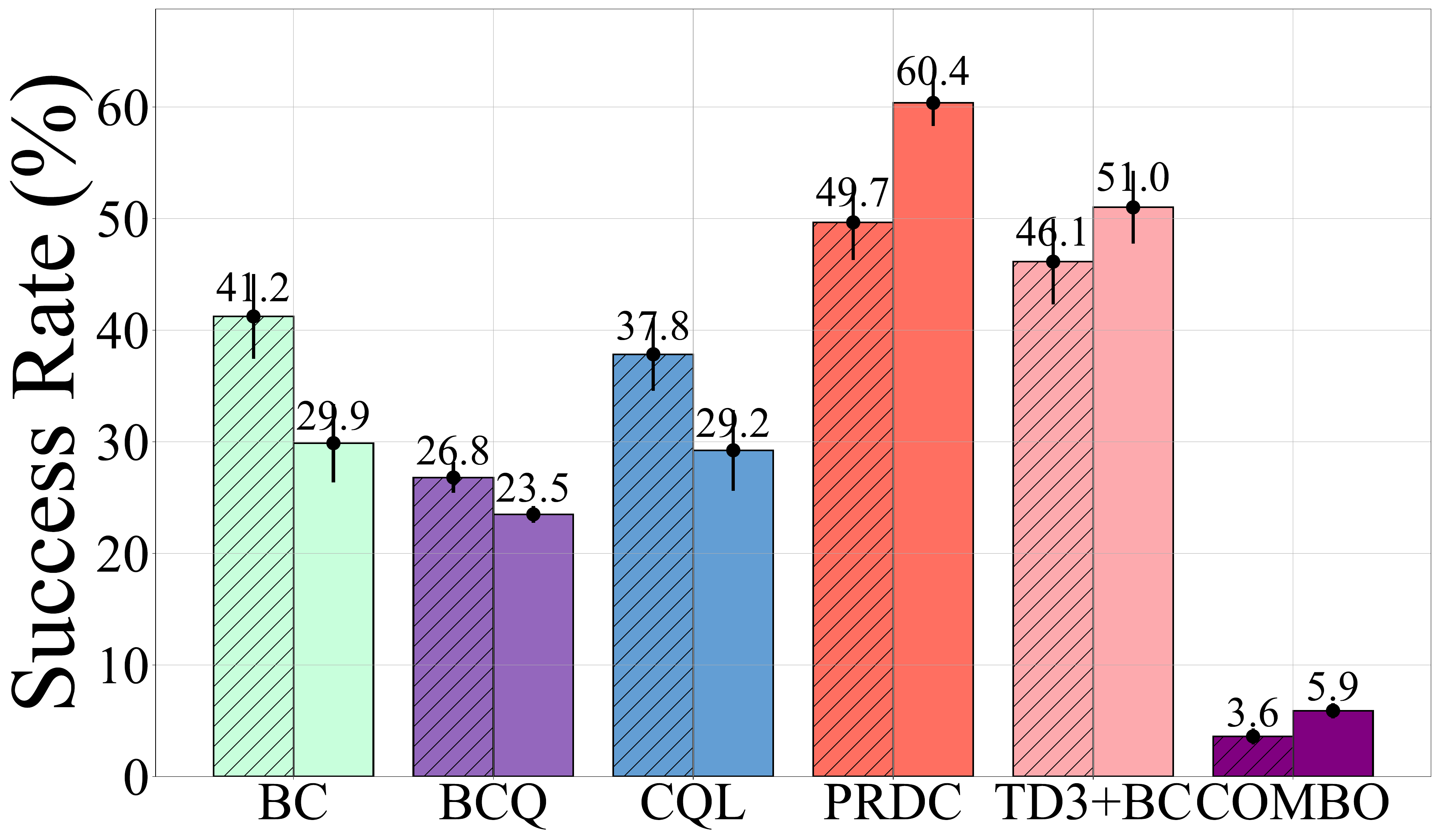}
}
\subfigure{
\includegraphics[width=0.225\textwidth]{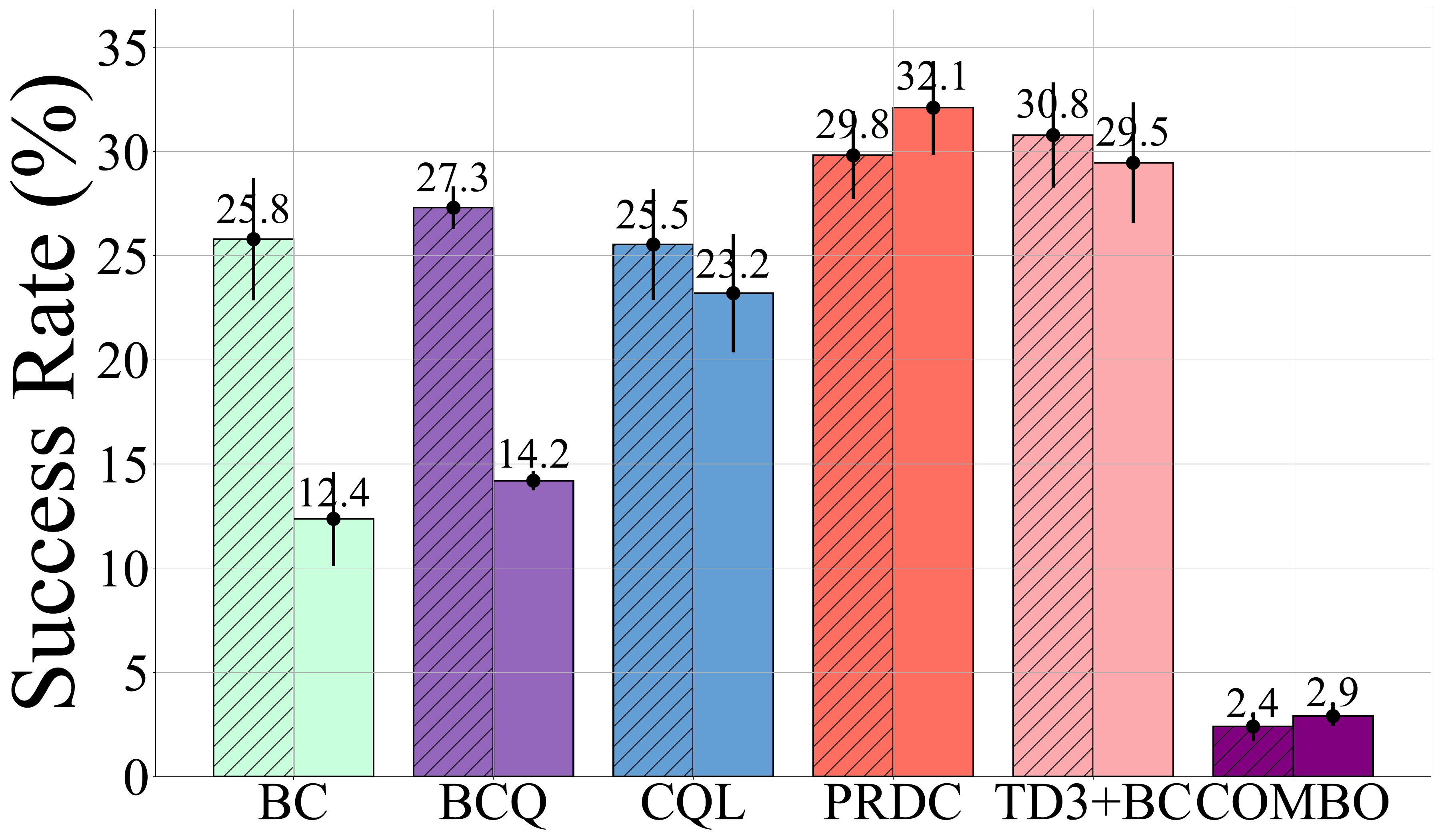}
}
\subfigure{
\includegraphics[width=0.225\textwidth]{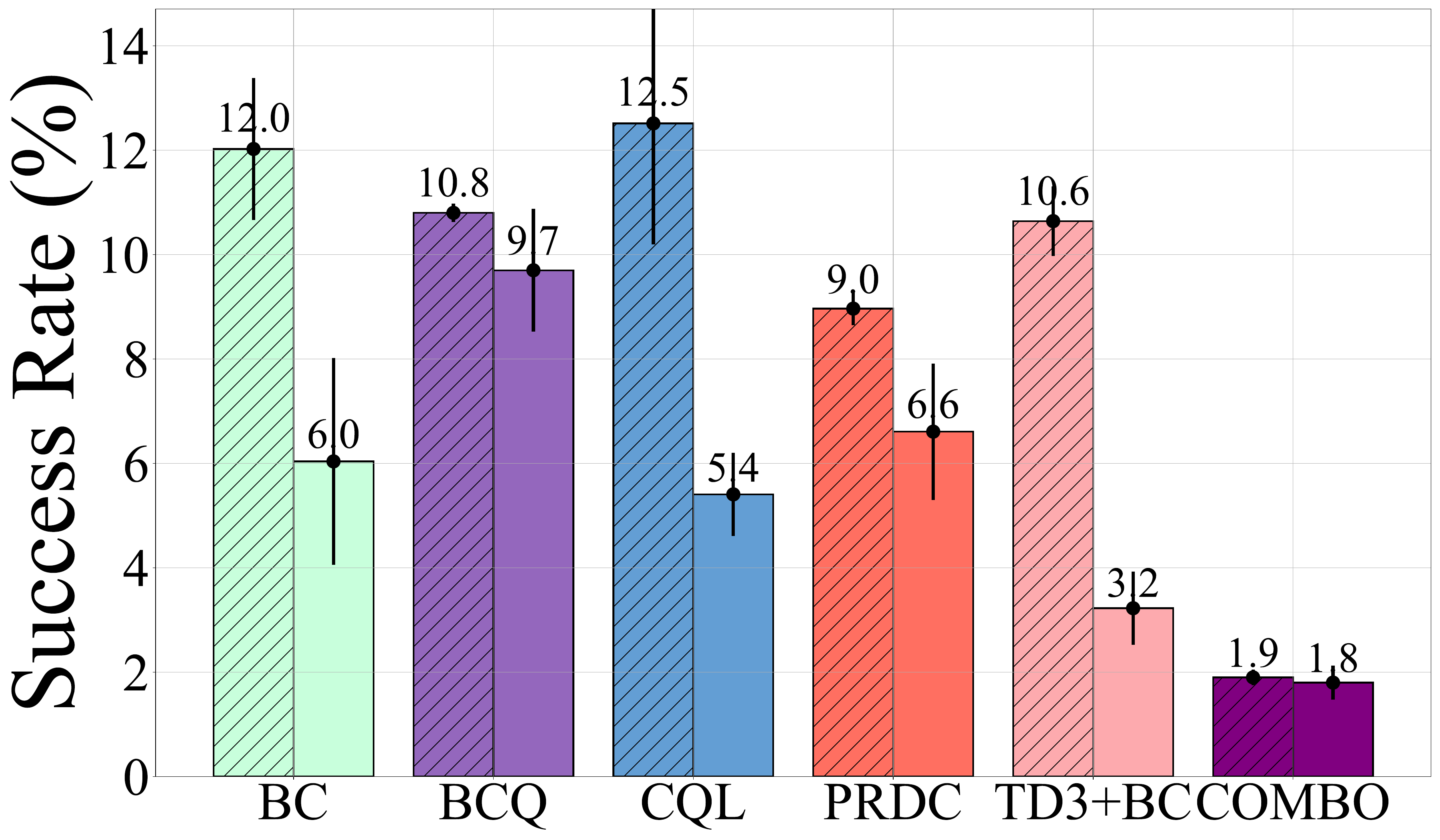}
}
\subfigure{
\includegraphics[width=0.225\textwidth]{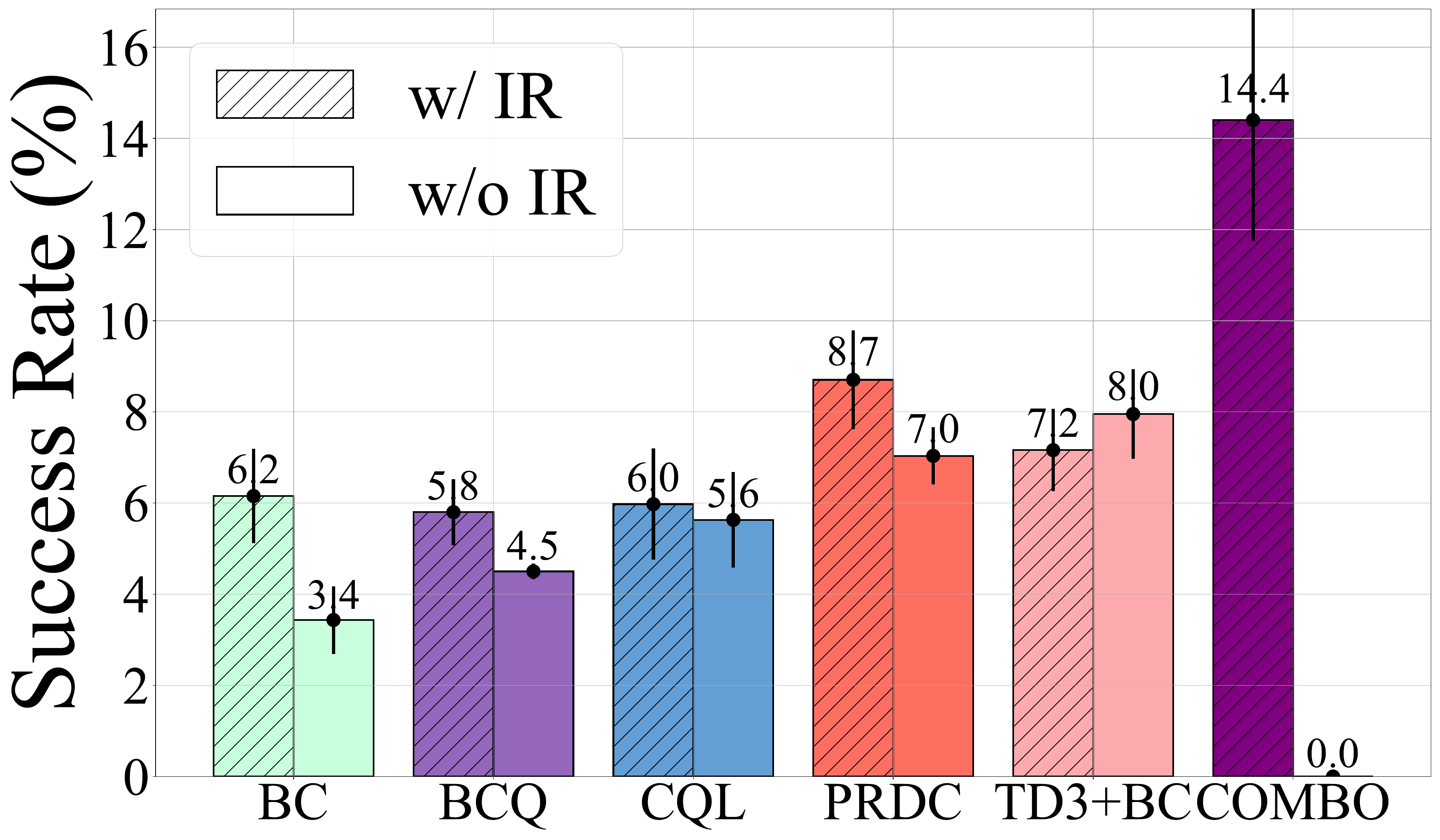}
} \\
\rotatebox{90}{\quad {\small CLEVR-Robot}}
\subfigure{
\includegraphics[width=0.225\textwidth]{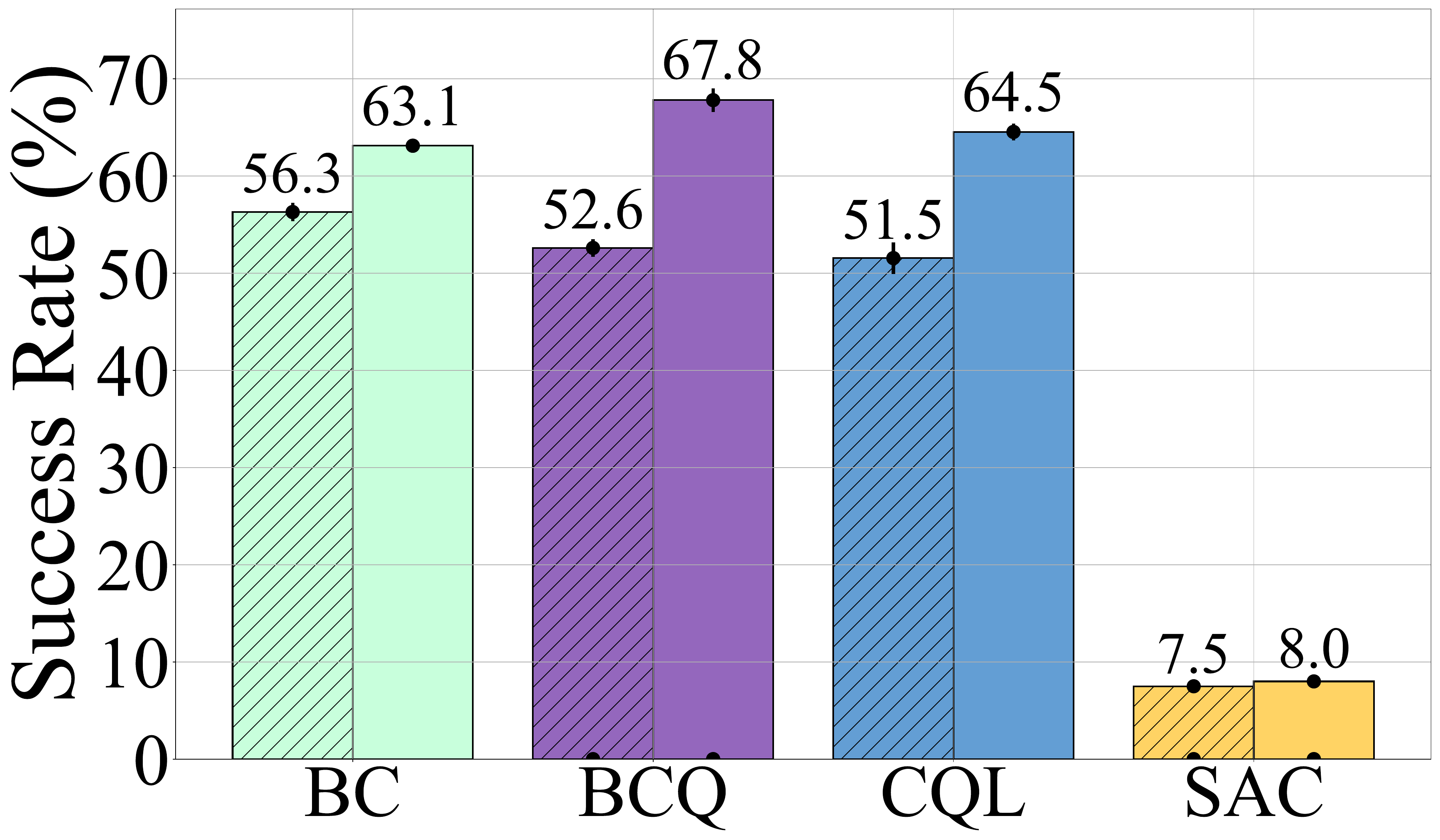}
\label{fig:task_in_od}
}
\subfigure{
\includegraphics[width=0.225\textwidth]{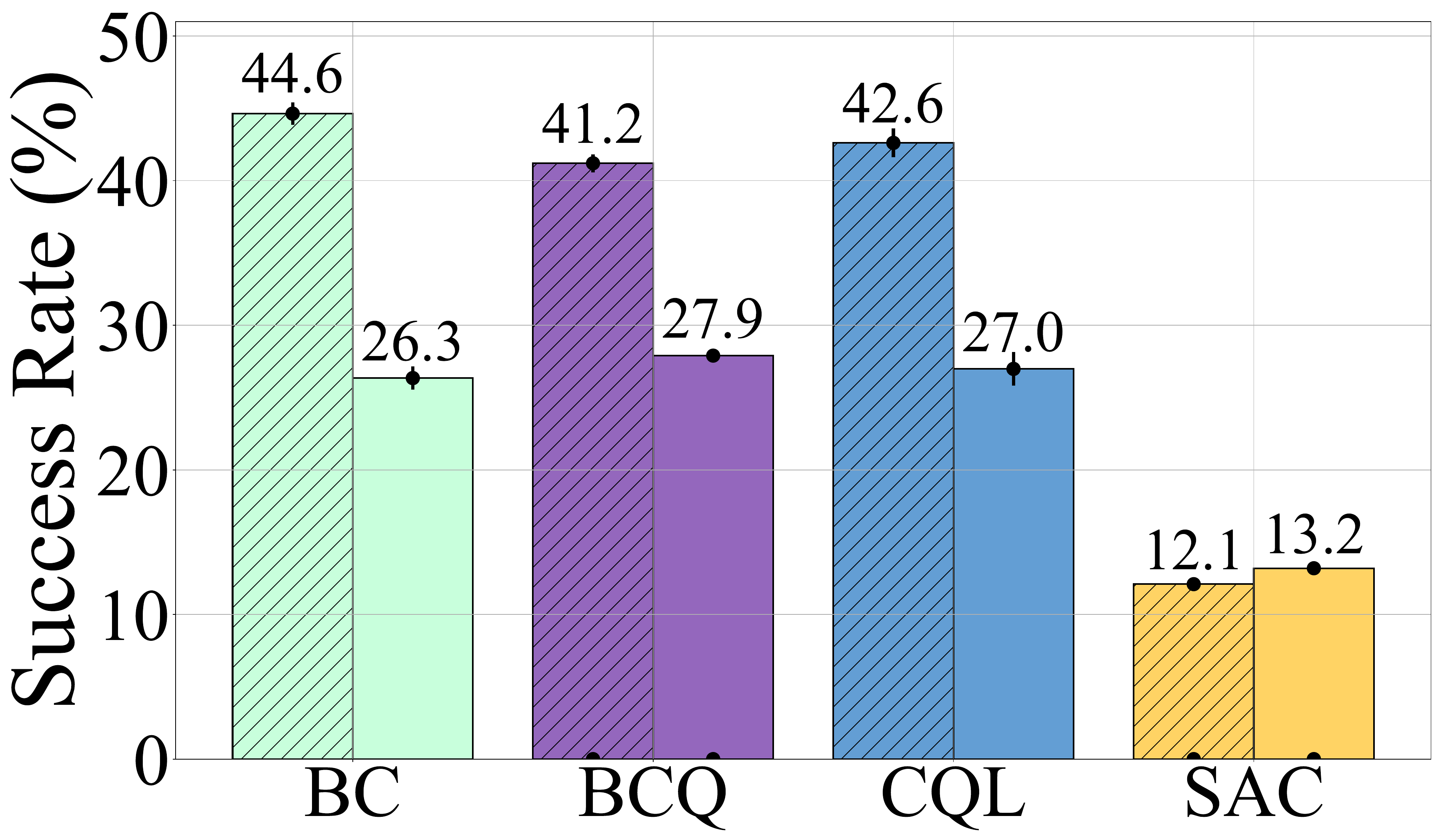}
}
\subfigure{
\includegraphics[width=0.225\textwidth]{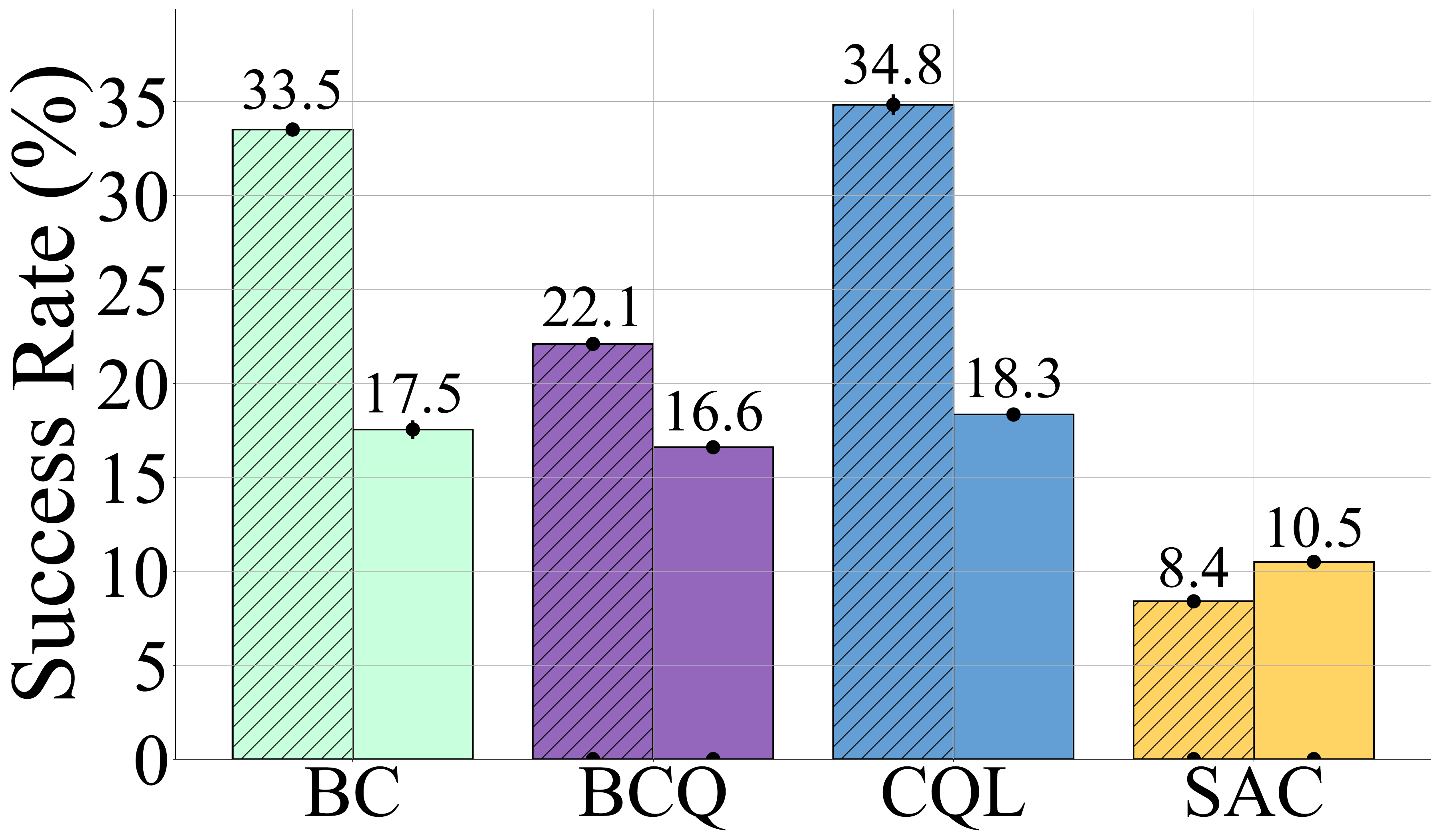}
}
\subfigure{
\includegraphics[width=0.225\textwidth]{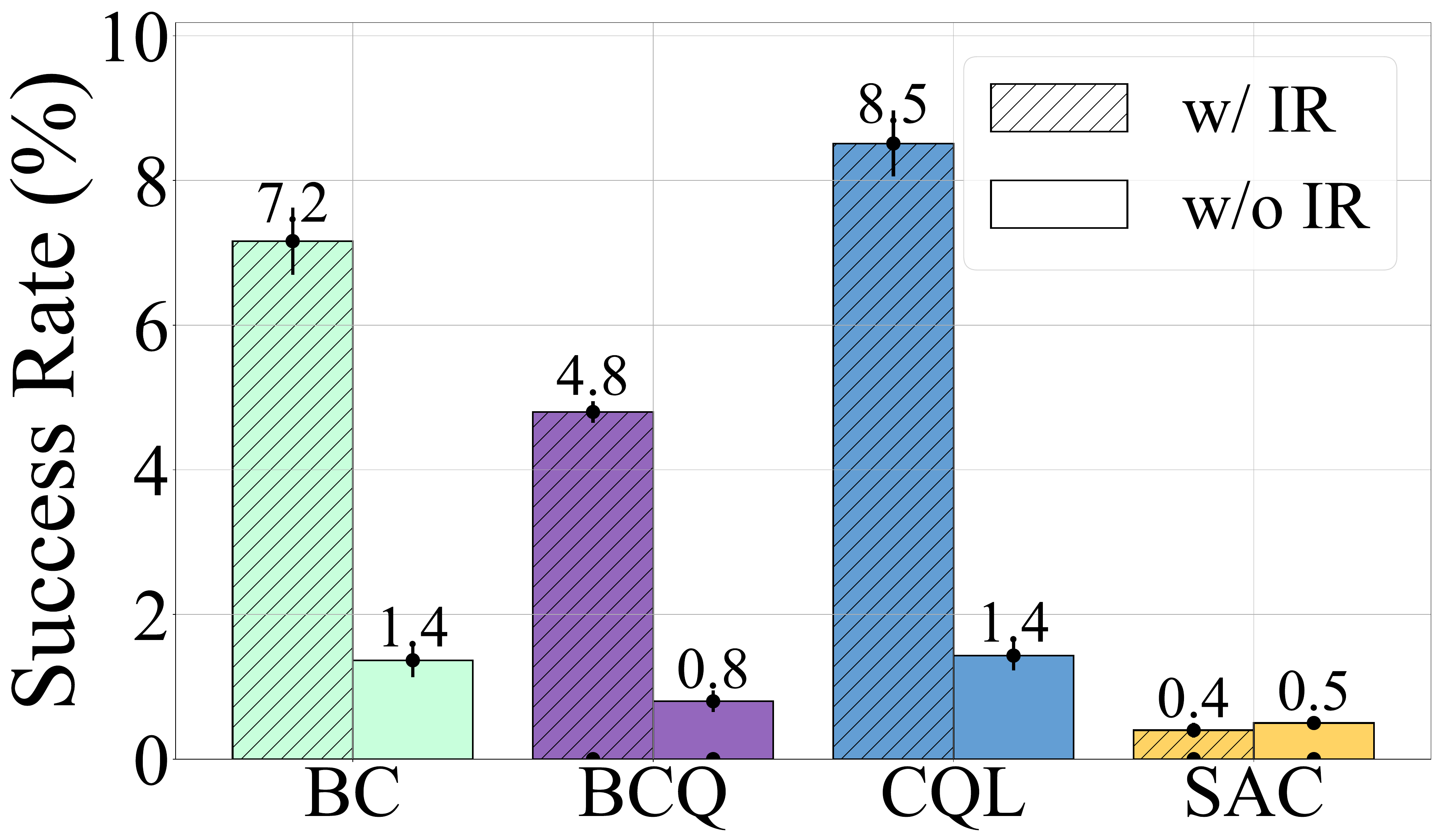}
} \\
\addtocounter{subfigure}{-12} %
\rotatebox{90}{\indent \ \indent \ \indent \ \indent BabyAI}
\subfigure[Training]{
\includegraphics[width=0.225\textwidth]{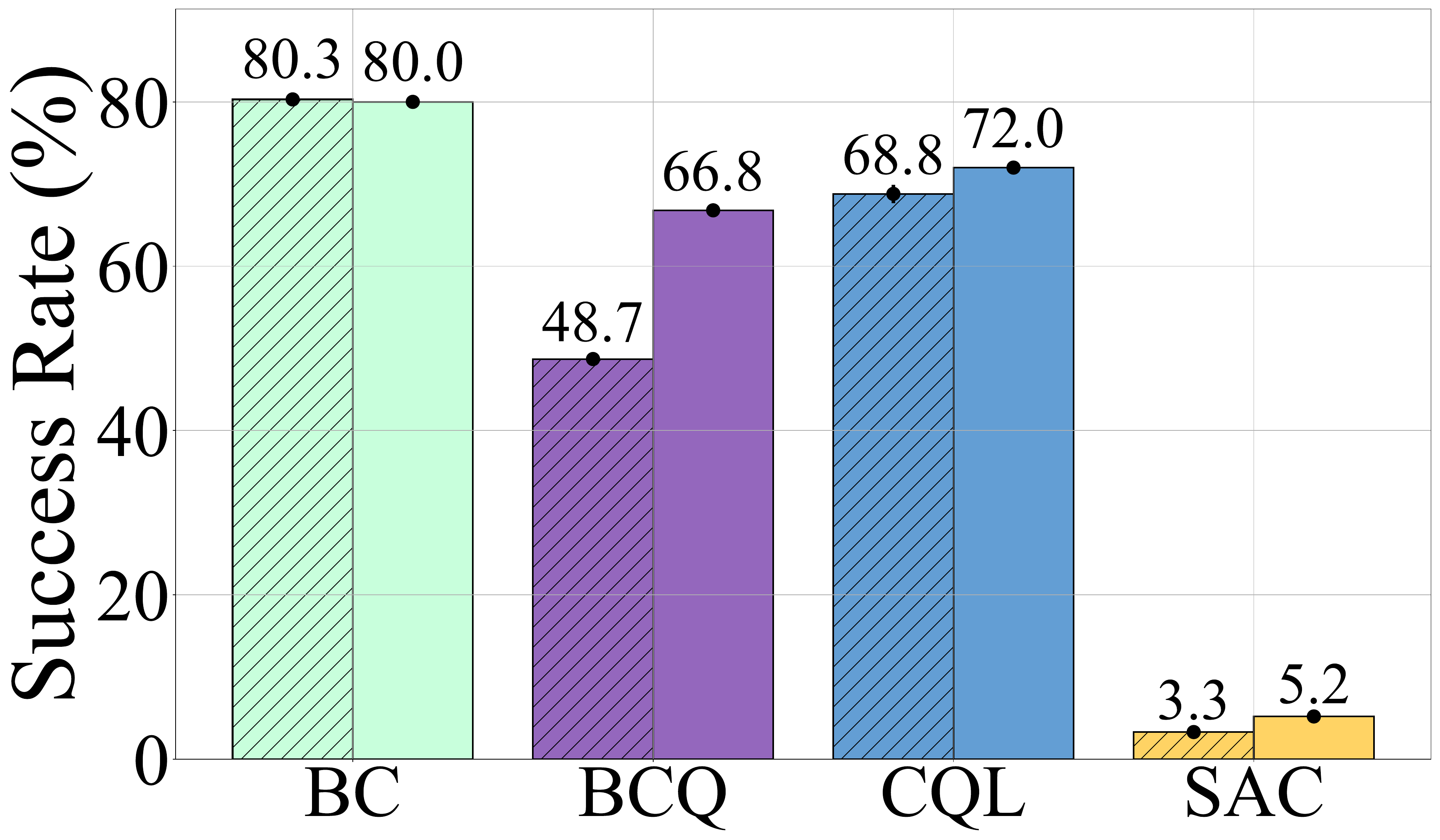}
}
\subfigure[Rephrasing]{
\includegraphics[width=0.225\textwidth]{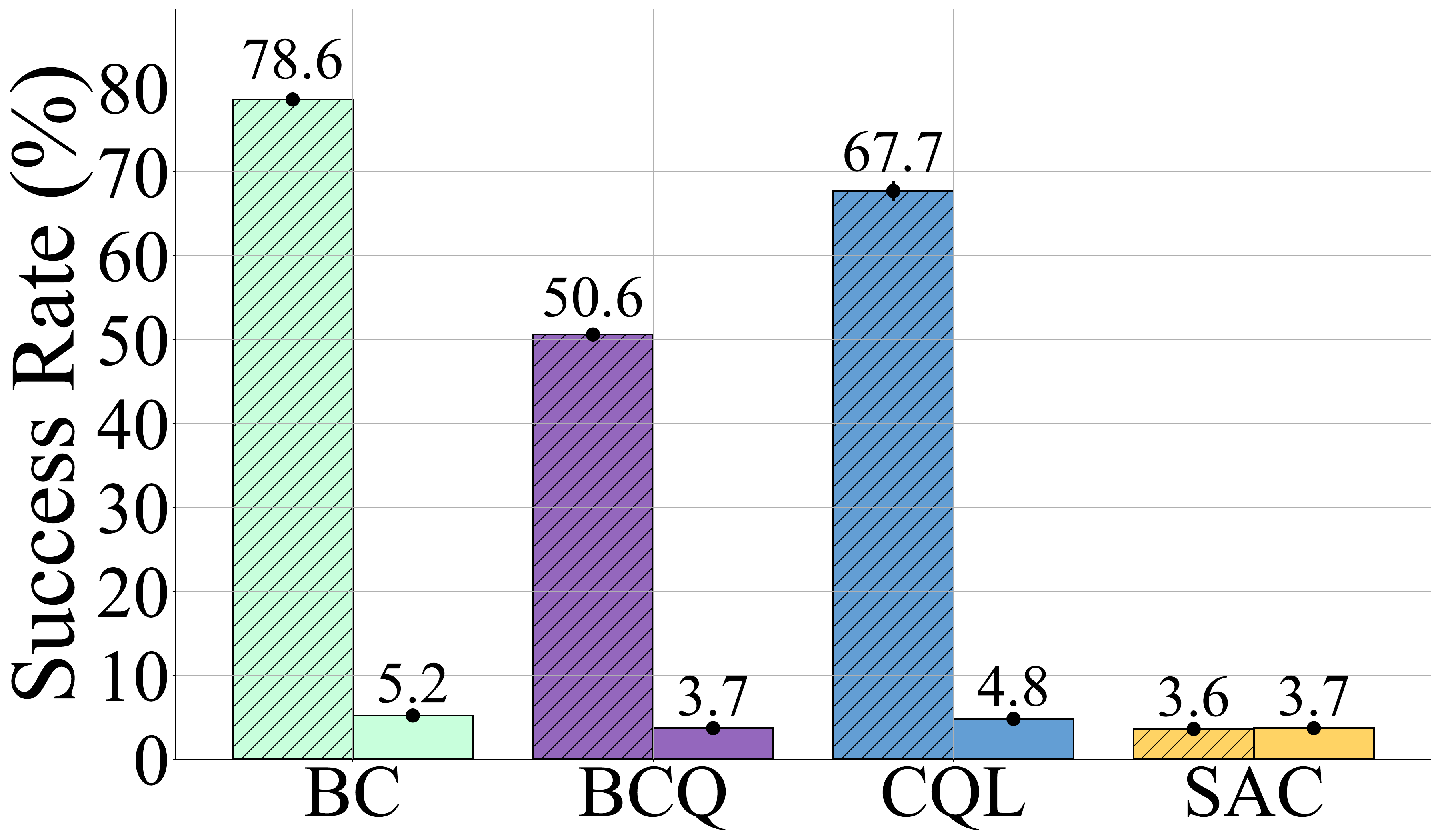}
}
\subfigure[Easy]{
\includegraphics[width=0.225\textwidth]{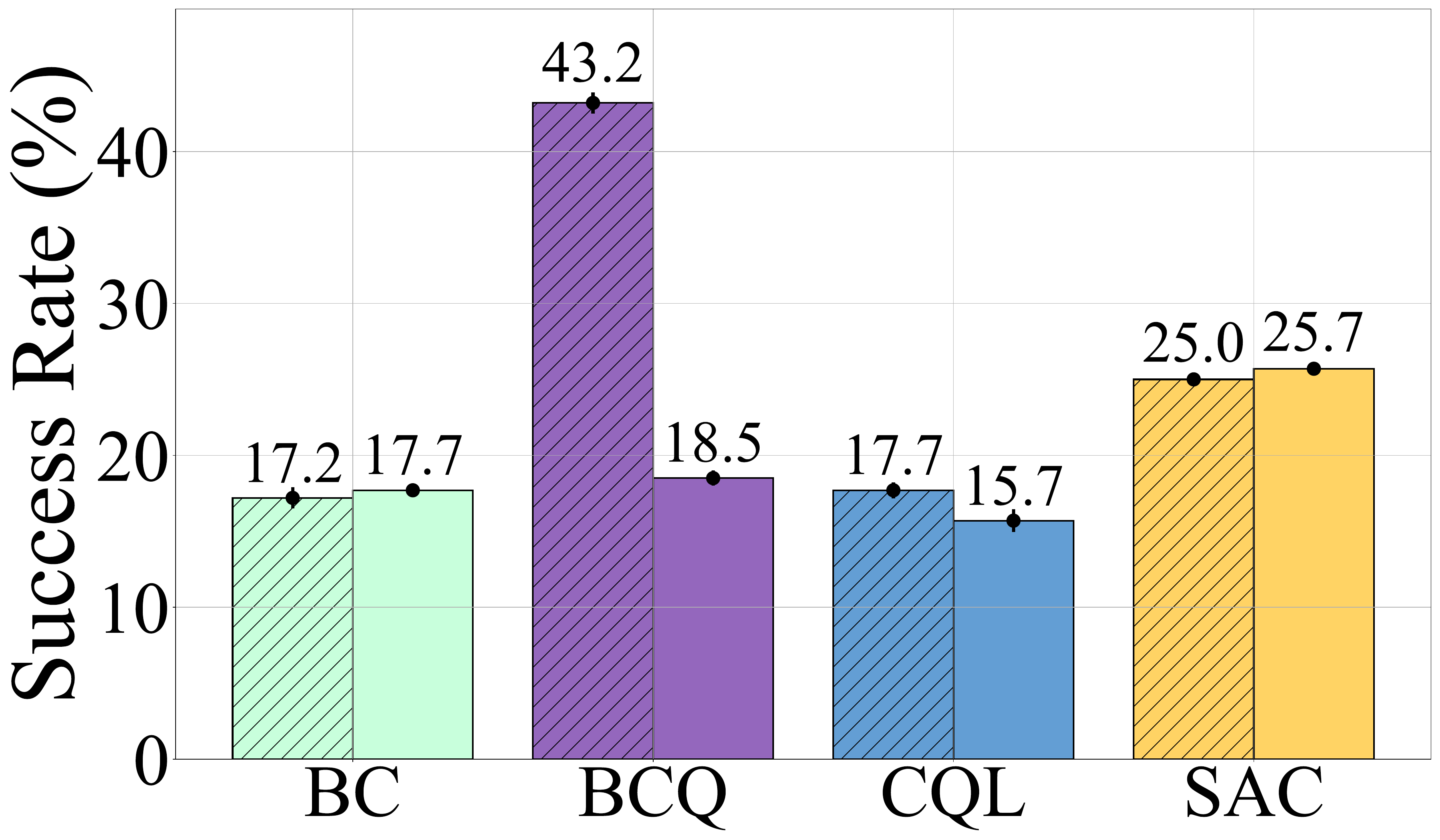}
}
\subfigure[Hard]{
\includegraphics[width=0.225\textwidth]{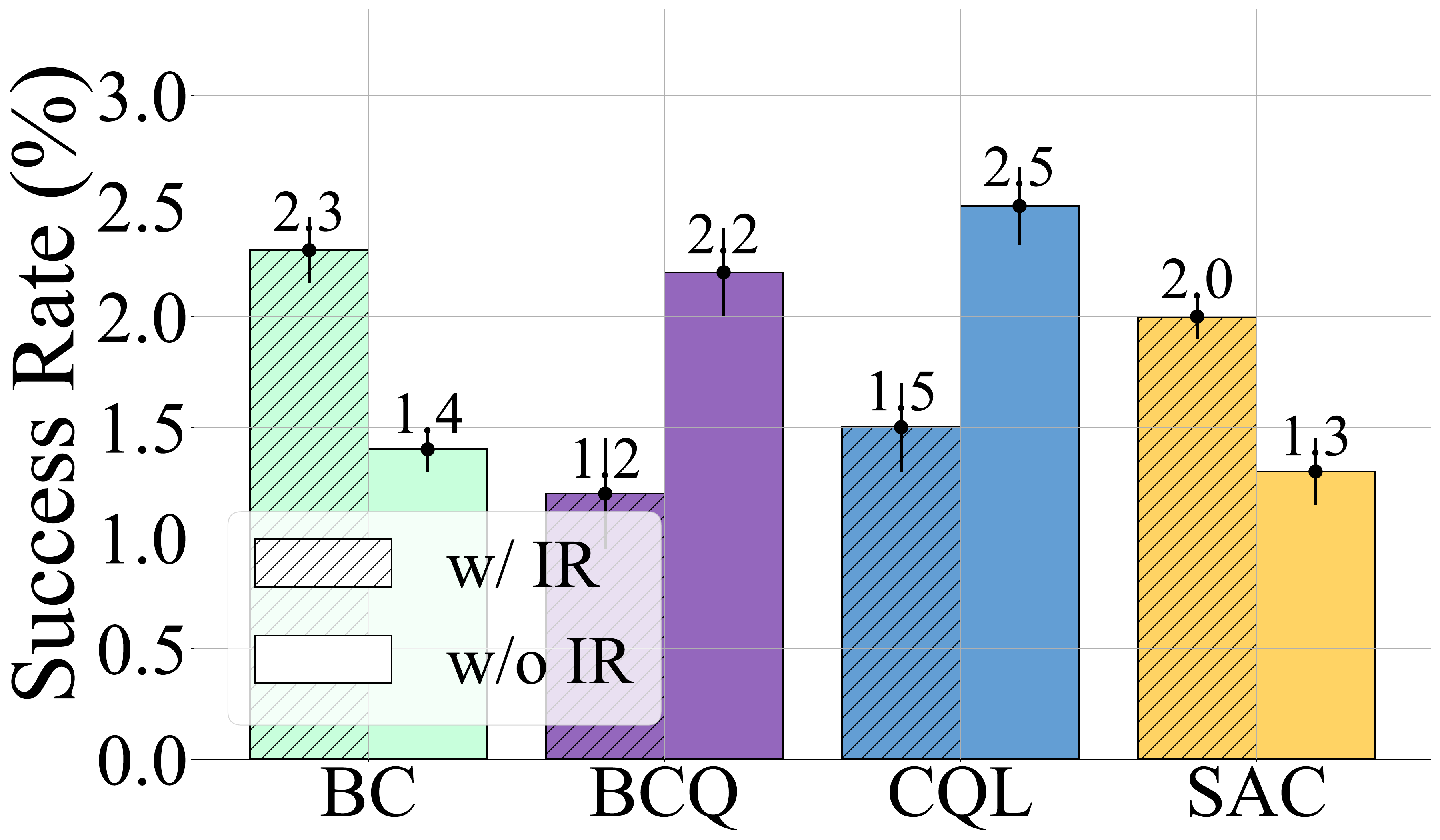}
}
\caption{Success rate bars of different methods on various levels of goals. The x-axis denotes the offline RL algorithm, and the y-axis denotes the success rate for completing various natural language goals. 'w/ IR' stands for training with imaginary rollouts. The success rate is calculated based on the average of the last five checkpoints, and the error bars stand for the half standard deviation over three random seeds.}
\label{fig:main_results}
\end{figure}

Fig. \ref{fig:main_results} presents the benchmark results of various offline RL trained with and without imaginary rollouts on \benchname~tasks. We have several main findings from the results.

\begin{itemize}[leftmargin=0.5cm]
    \item Performance improvement with LLM rollouts: Policies augmented with LLM-generated imaginary rollouts exhibit consistently higher performance on novel tasks than baseline methods. This suggests that LLM-based knowledge transfer enhances generalization and skill acquisition in unseen environments.
    \item Algorithmic comparisons: BC, CQL and BCQ outperform other offline RL algorithms across most tasks. BCQ and CQL achieve superior sample efficiency and stability in high-dimensional action spaces. As SAC is mainly used in an online setting, it fails to obtain high scores in our cases.
    \item Challenges in complex environments: Performance degrades significantly on hard tasks (e.g., BabyAI “Hard” levels), with success rates below 10\% on Meta-World, CLEVR-Robot, and BabyAI. This gap could stem from suboptimal reward function with current LLM rollouts, which may fail to encode task-specific constraints or long-horizon dependencies.
    \item Poor performance on LIBERO environment: All algorithms struggle with novel tasks on LIBERO due to its combinatorial complexity, indicating a need for advanced exploration strategies or hierarchical representations.
\end{itemize}

\subsection{Performance with Real Rollouts on Novel Tasks}
\label{sec:performance_with_real_novel}

\begin{figure}[h]
\centering
\subfigure[Rephrasing]{
\includegraphics[width=0.31\textwidth]{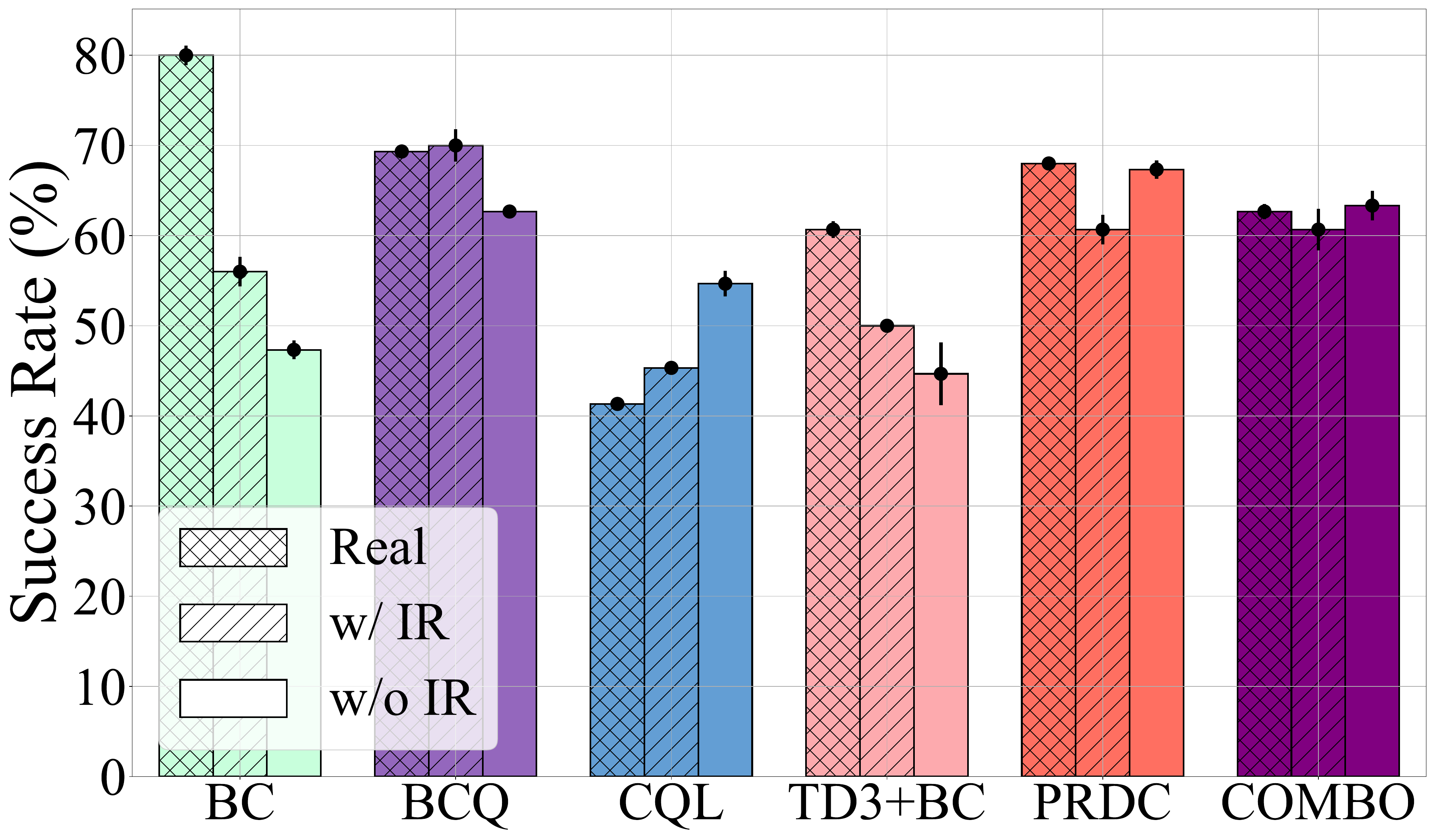}
}
\subfigure[Easy]{
\includegraphics[width=0.31\textwidth]{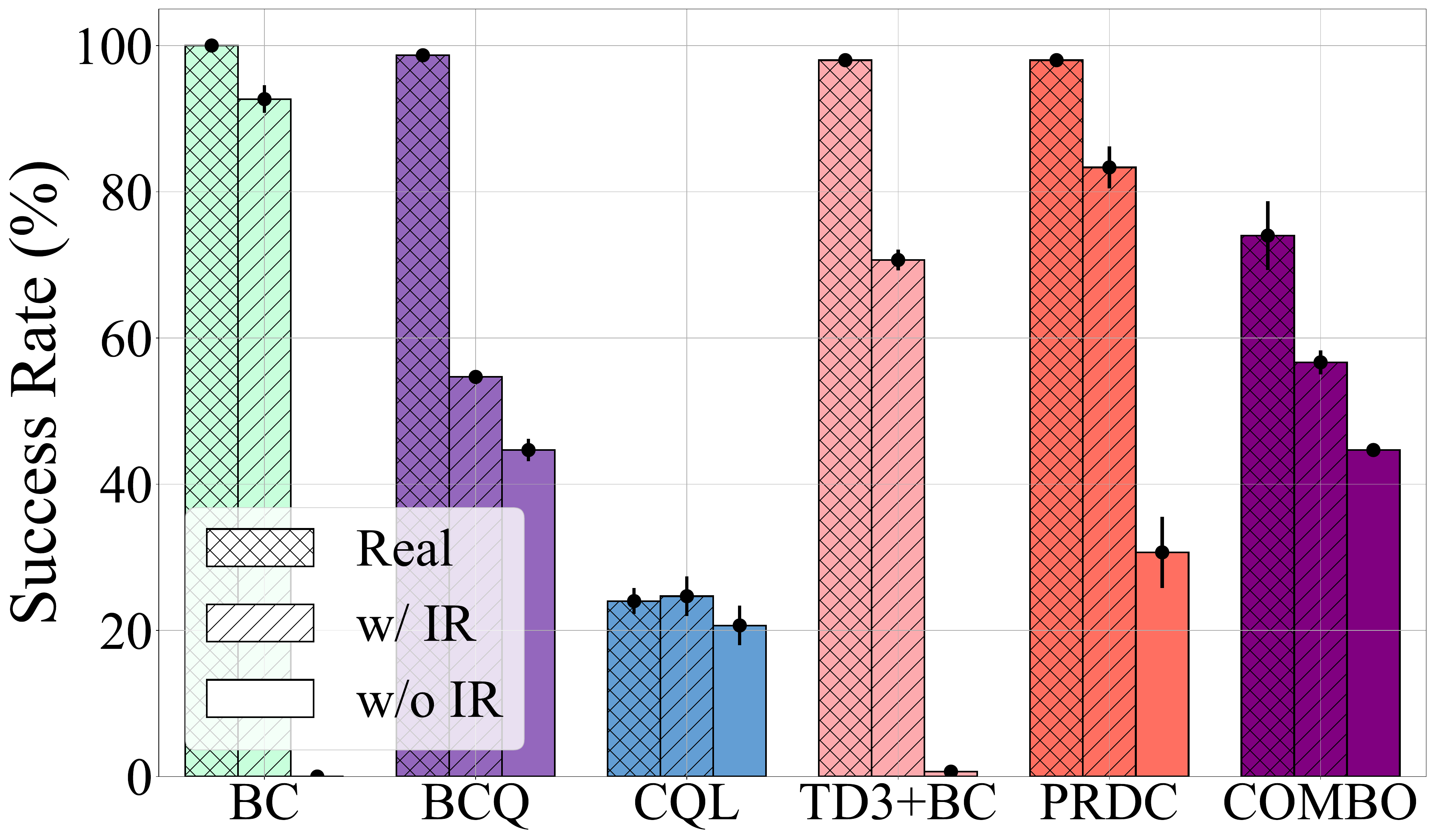}
}
\subfigure[Hard]{
\includegraphics[width=0.31\textwidth]{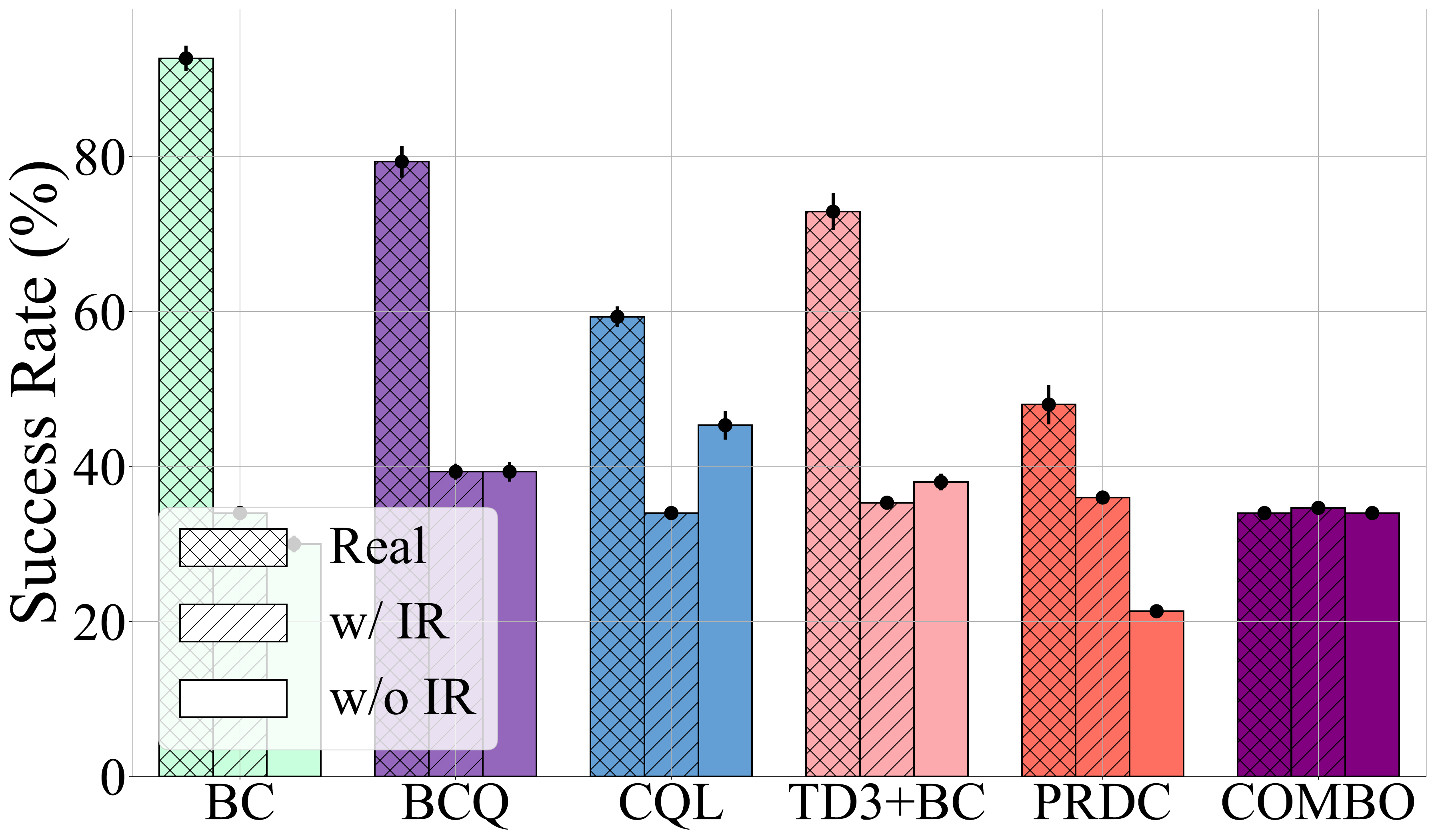}
}
\caption{Comparison of training with LLM-imaginary rollouts and real environmental rollouts on novel tasks. `Real' stands for the method trained with real environmental rollouts for novel tasks.}
\label{fig:real_novel}
\end{figure}

To investigate the improvement space for future algorithm development, we conduct experiments by training a policy with real rollouts on both training and novel tasks. Fig. \ref{fig:real_novel} shows the experiment results, with `Real' as the method trained on real rollouts of both training and novel tasks. In most of the tasks, Real outperforms or gets close performance to methods trained with IR, resulting in 64.37\% success rate for the Real method and 35.44\% for methods with IR in hard tasks. One exception is CQL on rephrasing task. This is attributed that the execution rollouts of the rephrasing task have already existed the dataset of real rollouts, with only the language expression of the instructions different. The conservative learning nature of CQL allows it to learn to focus on the features of the state, potentially enabling it performs well on rephrasing even when using only real rollouts for training tasks.

\section{Future Direction}

While demonstrating promising results for acquiring novel skills without online environment interactions, we suggest that RL with imaginary rollouts is still in the early stage of research and requires algorithmic development. We outline key directions for future research.

\textbf{Better offline RL algorithms for utilizing imaginary rollouts.} Current algorithms may not fully exploit the unique features of imaginary rollouts. Future work could focus on developing more advanced offline RL methods specifically handle the these imaginary rollouts. Since the quality and distribution of imaginary rollouts may differ from real ones, designing algorithms that can better account for the uncertainties and potential biases in imaginary data could lead to more effective policy learning. Besides, effectively filtering the low-quality or unreal rollouts in imaginary rollouts is an important direction.

\textbf{Unbiased and fast online adaptation and continual learning.} While RLIM reduces dependency on real-world interactions, practical deployment still requires efficient online adaptation to address imperfections in LLM imagination. A key challenge lies in avoiding catastrophic forgetting of pre-trained knowledge while rapidly fine-tuning policies with limited real interactions. Future research could focus on developing lightweight regularization techniques to preserve imaginary knowledge, meta-RL frameworks for few-shot adaptation, or progressive distillation methods to compress multi-task policies. Furthermore, designing bias correction mechanisms to disentangle inaccuracies in LLM-generated rollouts during online updates could enhance sample efficiency and stability.

\textbf{Better LLM imagination.} The quality of imaginary rollouts remains a limiting factor, as current LLMs often generate rollouts with insufficient grounding in physical or task-specific constraints. Future directions include improving LLM fine-tuning strategies, integrating physics-based simulators to validate generated rollouts, or developing iterative imagination processes where policy learning and LLM generation both get improvement. Additionally, scaling laws for LLM imagination—exploring how model size, prompt engineering, and domain-specific pretraining affect rollout quality—warrant systematic investigation.

\textbf{Vision-Language Models and Multi-Modal Imagination.} Current work mainly focuses on the numerical vectors as the observation. Extending RLIM to broader domains, e.g., vision, requires integrating vision-language models (VLMs) capable of processing and generating multi-modal rollouts. This entails addressing challenges such as aligning visual observations with language instructions, generating spatially consistent action sequences from pixel inputs, and handling partial observability in imagined states. Future work could explore cross-modal attention mechanisms for joint rollout generation, or develop hierarchical frameworks where high-level language plans guide low-level visual motion generation. 

\section{Conclusion}
\label{sec:conclusion}

In this work, we present \benchname, the first benchmark for RL with LLM-imaginary rollouts. By providing standardized datasets across locomotion, robotic manipulation, and navigation environments, \benchname~establishes a unified framework to evaluate offline RL algorithms that utilize the LLM-imaginary rollouts. The benchmark results reveal the limitations of existing offline RL methods when applied to hybrid real-imaginary datasets, underscoring the necessity for algorithmic innovations that better integrate LLM-generated knowledge. Beyond benchmarking, we hope to advance the research in agents that not only \textit{execute} predefined tasks, but \textit{imagine} solutions to novel challenges—a critical step toward generalizable embodied intelligence.

\clearpage

\bibliographystyle{apalike}
\bibliography{references}

\end{document}